\documentclass[lettersize,journal]{IEEEtran}
\usepackage{amsmath,amsfonts}
\usepackage{algorithmic}
\usepackage{algorithm}
\usepackage{array}
\usepackage[caption=false,font=normalsize,labelfont=sf,textfont=sf]{subfig}
\usepackage{textcomp}
\usepackage{stfloats}
\usepackage{url}
\usepackage{verbatim}
\usepackage{graphicx}
\usepackage{cite}
\usepackage{enumitem}
\usepackage{xcolor}
\usepackage{multirow}
\usepackage[table,xcdraw]{xcolor}
\newcommand{\redcell}[1]{\cellcolor[HTML]{F8EAF0}\textcolor{black}{#1}}
\newcommand{\bluecell}[1]{\cellcolor[HTML]{EAF8FD}\textcolor{black}{#1}}
\newcommand{\greencell}[1]{\cellcolor[HTML]{EEF2F5}\textcolor{black}{#1}}

\DeclareRobustCommand{\redmark}[1]{\colorbox[HTML]{F8EAF0}{\strut\textcolor{black}{#1}}}
\DeclareRobustCommand{\bluemark}[1]{\colorbox[HTML]{EAF8FD}{\strut\textcolor{black}{#1}}}
\DeclareRobustCommand{\greenmark}[1]{\colorbox[HTML]{EEF2F5}{\strut\textcolor{black}{#1}}}

\begin{document}

\title{Efficient Multi-modal Video Fusion Network with Dynamic Region-Guided Temporal Interaction}

\author{Jingxue Huang, Xiaosong Li, Yi Lin
\thanks{
Jingxue Huang and Yi Lin are with the College of Computer Science, Sichuan University, Chengdu 610000, China. (e-mail: jasonwong30@163.com; yilin@scu.edu.cn)

Xiaosong Li is with the Guangdong-Hong Kong-Macao Joint Laboratory for Intelligent Micro-Nano Optoelectronic Technology, School of Physics and Optoelectronic Engineering, Foshan University, Foshan 528225, China. (e-mail: lixiaosong@buaa.edu.cn)

}
}

\markboth{Journal of \LaTeX\ Class Files,~Vol.~X, No.~X, August~2026}%
{Shell \MakeLowercase{\textit{et al.}}: A Sample Article Using IEEEtran.cls for IEEE Journals}


\maketitle

\begin{abstract}
Multi-modal video fusion aims to generate temporally coherent videos with salient targets and rich structural details by integrating complementary information from heterogeneous sensors. However, existing video fusion methods commonly rely on optical flow estimation and feature warping for inter-frame alignment and cross-frame information aggregation to maintain temporal consistency, thereby introducing considerable computational overhead and making the fusion performance vulnerable to flow estimation errors. To address these limitations, this paper proposes an efficient multi-modal video fusion network with dynamic region-guided temporal interaction. Specifically, the proposed spatio-temporal interaction module leverages the Test-Time Training paradigm to model temporal dependencies with linear complexity. Building on this design, the module restricts temporal modeling to key dynamic regions in the video sequence, as static background regions provide limited temporal cues. By excluding low-variation tokens from unnecessary temporal interaction, this selective strategy contributes to improved fusion quality and temporal consistency. In addition, leveraging the strong ability of pretrained MAE features to capture global structural dependencies, we introduce a feature-level temporal consistency constraint to suppress local flickering by aligning the inter-frame variation directions of the fused and source videos. We further propose a new MAE-based temporal metric that quantifies directional distance between their temporal variations within dynamic regions. Experimental results demonstrate that, compared with existing state-of-the-art methods, our method achieves superior fusion quality and temporal consistency on multiple benchmark datasets, while exhibiting high computational efficiency under edge-device deployment settings.
\end{abstract}

\begin{IEEEkeywords}
Multi-modal video fusion, dynamic regions, sequence modeling, feature-level temporal consistency constraint, temporal metric.
\end{IEEEkeywords}

\section{Introduction}
\IEEEPARstart{D}{ue} to the inherent limitations of modality-specific imaging mechanisms, a single modality is generally unable to provide a comprehensive representation of a scene~\cite{VideoFusion}. Specifically, visible sensors provide rich texture, color, and structural details, which are beneficial for preserving the natural visual representation of scenes. However, their imaging quality is highly susceptible to illumination variations, nighttime conditions, smoke, and adverse weather. In contrast, infrared sensors capture thermal radiation from targets, offering stronger target perception under low-light and other degraded conditions. Nevertheless, infrared imaging often suffers from insufficient texture details and relatively low spatial resolution~\cite{TIP2,TIP4}. Multi-modal video fusion (MMVF) jointly exploits complementary information from video sequences captured by heterogeneous sensors to generate unified and information-rich video representations~\cite{VF-Bench,temcoco,EvaNet,ULMGFuse}. This capability supports reliable scene understanding in complex environments, facilitating applications such as video surveillance, autonomous driving, low-altitude observation, and disaster detection~\cite{TIP1,TIP5,TIP6,mask-dif,zhangjian1}.

\begin{figure*}[t]
   \centering
   \includegraphics[width=0.85\linewidth]{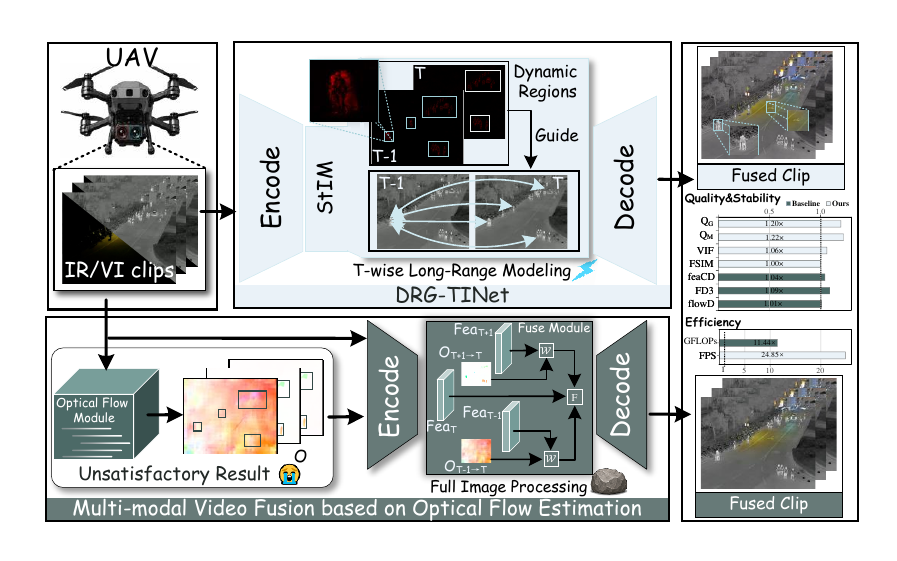}   
   \vspace{-20pt}
   \caption{Comparison with an existing video fusion method~\cite{VF-Bench} in terms of temporal interaction schemes, fusion quality and computational efficiency. All reported metrics are evaluated on the VTMOT dataset.}
   \label{fig1}   
\end{figure*}

Existing advanced video fusion methods commonly resort to optical-flow estimation and explicit feature warping to achieve cross-frame information modeling and rigid temporal alignment~\cite{mambavf}. For example, UniVF~\cite{VF-Bench} introduces such techniques into multi-frame learning to establish inter-frame relationships and achieves promising video fusion performance. However, as illustrated in Fig.~\ref{fig1}, flow-based methods suffer from two fundamental bottlenecks. The first one lies in their low computational efficiency, while the second is their high sensitivity to estimation accuracy. Specifically, the fusion quality is strictly constrained by the accuracy and stability of optical-flow estimation. In complex dynamic scenarios, optical flow is often difficult to estimate reliably, and the accumulated errors over long video sequences can be further propagated into the fusion method. In addition, if optical flow is discarded and global self-attention-based Transformers are adopted for direct spatio-temporal information aggregation, even heavier computational overhead will be introduced. Because establishing correlations along any dimension with conventional global attention incurs quadratic complexity, the computational cost grows prohibitively with the sequence length, making it difficult to scale to long video sequences. These issues severely hinder the practical deployment of MMVF on resource-constrained edge devices, where real-time inference and low power consumption are typically required.

To address these limitations, this work aims to explore a lightweight inter-frame relation modeling strategy that efficiently models temporal dependencies without relying on explicit optical-flow alignment or dense global spatio-temporal attention. Recently, Test-Time Training (TTT)~\cite{TTT} has shown strong potential for efficient visual sequence modeling. By reformulating token interaction as an online learning process over a compact inner model, TTT avoids explicitly constructing pairwise attention maps and enables sequence modeling with linear computational complexity. Although existing visual TTT architectures have demonstrated the effectiveness of TTT in modeling intra-frame spatial context, the use of TTT-based sequence modeling for efficient cross-frame interaction in MMVF remains underexplored.

In this paper, we propose an efficient MMVF network equipped with a spatio-temporal interaction module (STIM) for lightweight temporal dynamic relation modeling. Specifically, STIM employs TTT-based sequence modeling along the temporal dimension to efficiently capture cross-frame dependencies. Nevertheless, uniformly applying temporal interaction to all inter-frame features would still introduce substantial redundant feature interactions, as different spatial regions provide unequal temporal cues. Most static background regions exhibit only minor variations across adjacent frames, and modeling these regions may provide limited temporal benefit while propagating unnecessary noise. In contrast, moving targets, regions with changing boundaries, and areas exhibiting distinct cross-modal response differences are more likely to cause temporal discontinuities in the fused video. Therefore, as shown in Fig.~\ref{fig1}, we introduce a dynamic region guidance mechanism into STIM, allowing it to gather the dynamic features from each frame, reorganize them into a sequence of dynamic features, and perform TTT-based temporal interaction over the sequence. In this way, each selected feature can more effectively aggregate inter-frame context from the video clip, enabling sparse temporal modeling focused on dynamically informative regions. In addition, to further enhance the temporal stability of fused videos, we introduce a feature-level temporal consistency constraint based on pretrained MAE representations. Benefiting from their strong ability to capture global structural dependencies, MAE features provide a structure-aware representation space in which patch-level structural and regional variations between frames can be effectively characterized. We therefore align the adjacent-frame feature variations of the fused video with those of the source videos. The main contributions of this work are summarized as follows:
\begin{itemize}[label=-, leftmargin=*, itemsep=2pt, topsep=1pt]
     \item We propose an efficient MMVF framework that employs linear-complexity TTT-based sequence modeling for temporal interaction and dynamic-region guidance to further focus inter-frame modeling on dynamically informative content.
    
     \item We propose a feature-level temporal consistency constraint that exploits MAE representations to align the temporal variations of fused videos with those of source videos, thereby suppressing inter-frame flickering in MMVF.
    
     \item We introduce a new feature-level temporal metric, termed Feature-Level Dynamic Direction Distance (FD3), to evaluate temporal consistency from the perspective of region-level dynamic variations in the feature space.
    
     \item Extensive experiments on multiple benchmark datasets demonstrate that our method achieves superior fusion quality. We further validate that the proposed method exhibits higher computational efficiency than the compared methods when deployed on the Jetson AGX Orin platform.
\end{itemize}

\section{Related Work}
\subsection{Multi-modal Image Fusion}
With the rapid advancement of deep learning, multi-modal image fusion (MMIF) has progressively evolved from conventional paradigms relying on handcrafted rules and traditional image-processing operators~\cite{lixilaitip,xiaosongli1} toward data-driven learning paradigms centered on neural network architectures~\cite{UPFusion,lixilaitip2,mma}, leading to substantial improvements in overall fusion quality. According to the underlying network architectures~\cite{tpamireview25}, existing MMIF methods can be broadly categorized into convolutional neural network (CNN)-based~\cite{LRRNet,DeDNet,CNN1}, Transformer-based~\cite{wavefusion,Transformer1,Frefusion}, generative adversarial network (GAN)-based~\cite{PRRGAN,DDCGAN,GAN1}, diffusion model-based~\cite{shiyuCVPR26,mask-dif,Diffusion1}, Mamba-based~\cite{shufflemamba,mamba1,mamba2}, and hybrid-architecture methods~\cite{ULMGFuse,Huangjingxue1,xia2026rethinking}. For CNN-based methods, Li et al.~\cite{LRRNet} proposed LRRNet, which unfolds the optimization process of low-rank representation into a lightweight feed-forward convolutional network, thereby reducing reliance on empirical architecture design. For Transformer-based methods, Wu et al.~\cite{Transformer1} developed FCFormer based on a generalized fully-connected self-attention mechanism. It exploits both local and non-local correlations among multiple source images and provides a unified framework for different image fusion tasks. For GAN-based methods, Huang et al.~\cite{PRRGAN} introduced an end-to-end framework for trimodal medical image fusion. The method mainly incorporates primitive relationship reasoning to enhance global semantic information.  For diffusion-based methods, Shi et al.~\cite{shiyuCVPR26} proposed a degradation-aware diffusion framework for MMIF under arbitrary degradation conditions. The model directly regresses the fused image for implicit denoising and introduces a joint observation-model correction mechanism to enforce both degradation and fusion constraints during sampling. For Mamba-based methods, Cao et al.~\cite{shufflemamba} proposed Shuffle Mamba, which adopts random and inverse shuffle operations to alleviate the scanning bias caused by fixed token-ordering strategies. The framework further performs modality-aware representation and cross-modal interaction along both spatial and channel dimensions.  For hybrid architectures, Xia et al.~\cite{xia2026rethinking} revisited the integration of CNNs and Transformers from a frequency perspective, aiming to better coordinate the local-detail modeling capability of CNNs with the global-context modeling capability of Transformers.

\subsection{Multi-modal Video Fusion}
To address the temporal instability of frame-wise MMIF in dynamic scenes, recent studies have shifted their attention to MMVF. Gong et al.~\cite{temcoco} proposed an infrared–visible video fusion framework, which incorporates temporal cooperative modules and targeted guidance to achieve superior fusion results. Tang et al.~\cite{VideoFusion} proposed a unified multi-modal video fusion and restoration framework with bi-temporal co-attention, which exploits temporal and cross-modal complementary cues for information restoration and fusion. Although these methods improve temporal coherence through strong spatio-temporal modeling, heavy multi-frame aggregation may over-couple features under complex motion or large temporal variations, causing detail loss and inter-frame artifacts. Zhao et al.~\cite{VF-Bench} introduced a unified video fusion framework, employing an additional optical flow network to estimate inter-frame pixel displacement and perform warping. Although these MMVF methods achieve superior fusion quality, their complex temporal interaction or warping mechanisms often incur substantial computational overhead as well as high inference latency, limiting real-time deployment on resource-constrained edge devices. In this paper, the proposed method employs TTT-based sequence modeling along the temporal dimension and performs inter-frame interaction within dynamic regions, without relying on explicit optical-flow alignment or heavy multi-frame aggregation.

\subsection{Efficient Visual Sequence Modeling} The Vision Transformer (ViT) architecture~\cite{VIT} has been widely adopted in image fusion with remarkable performance. However, standard Softmax attention explicitly computes pairwise interactions between queries and keys, resulting in a computational complexity of $\mathcal{O}(N^{2})$ with respect to the sequence length. Consequently, its computational cost increases substantially when processing long visual sequences. To alleviate this computational bottleneck, linear attention has emerged as an alternative owing to its $\mathcal{O}(N)$ complexity~\cite{linearattention}. Specifically, it replaces the Softmax operation with a linear kernel mapping and reorganizes the computation order from $(QK^{\top})V$ to $Q(K^{\top}V)$, thereby avoiding the construction of a complete attention matrix. Despite its considerable efficiency gains, aggregating the key--value information into a shared context $K^{\top}V$ restricts nonlinear relation modeling and fine-grained query--key interactions, leading to limited representational capacity~\cite{han2023flatten,qincosformer}. To better balance computational efficiency and representation capability, Test-Time Training reformulates attention as an online learning process, in which a compact nonlinear internal model is updated at inference time using key--value pairs, thereby enabling adaptive token--context interactions with $\mathcal{O}(N)$ complexity~\cite{VITTT}. 

\begin{figure*}[t]
   \centering
   \includegraphics[width=0.85\linewidth]{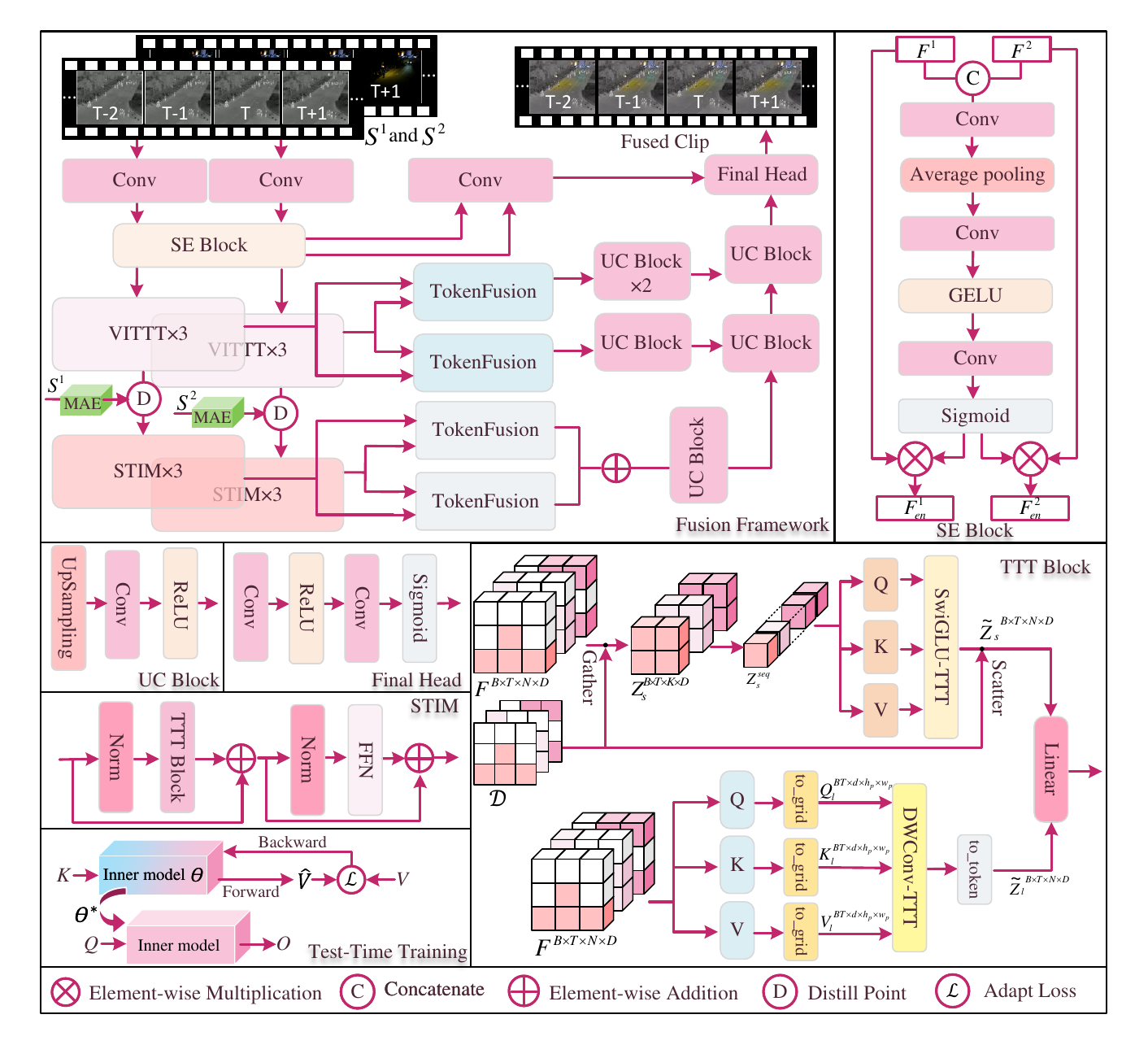}   
   \vspace{-15pt}
   \caption{Overview of the proposed multi-modal video fusion framework.}   
   \label{fig2}   
\end{figure*}

\section{Methodology}
\subsection{Overall Framework}
Given paired multi-modal video sequences $S^1, S^2 \in \mathbb{R}^{B \times T \times 3 \times H \times W}$, where $B$, $T$, $H$, and $W$ denote the batch size, number of frames, height, and width, respectively, the proposed method aims to generate high-quality fused results $S^F$ with high computational efficiency. As illustrated in Fig.~\ref{fig2}, the overall framework is organized into three stages.

First, in the encoder, the input sequences are sequentially fed into convolutional layers and Squeeze-and-Excitation (SE) Blocks~\cite{SEBlock} to perform shallow feature extraction and mutual feature enhancement. Since convolutional operations are inherently limited by their local receptive fields, modeling global contextual dependencies is essential for preserving spatial structures. Therefore, the enhanced features are further delivered to VITTT~\cite{VITTT} to capture intra-frame long-range dependencies. Compared with conventional vision Transformers with quadratic self-attention complexity, VITTT enables global context modeling with linear computational complexity, making it well suited for efficient MMVF. Subsequently, the extracted features are fed into the proposed STIM to establish inter-frame temporal dependencies. 

Second, in the fusion stage, the features extracted by VITTT and STIM are fused through TokenFusion, where a lightweight fully connected layer is employed to integrate multi-modal features. Specifically, the outputs of the first and third VITTT blocks, together with those of the second and third STIM blocks, are selected as inputs to TokenFusion.

Finally, in the decoder, the fused features are progressively upsampled and reconstructed. Meanwhile, the enhanced shallow features extracted by the SE Blocks are integrated and introduced into the final head through a residual connection to facilitate the final reconstruction, producing the fused sequences.

\subsection{Spatio-Temporal Interaction Module}
Modeling temporal dependencies across consecutive frames is essential for maintaining temporal consistency in fused videos. To this end, we propose the STIM, which consists of two key components: dynamic region estimation and efficient spatio-temporal interaction. Specifically, a dynamic region map is generated for each frame to identify regions that are most relevant to temporal variation. In addition, temporal interactions are performed within these dynamic regions. Since both components are subsequently implemented at the patch level, each frame is divided into $N$ non-overlapping patches of size $p\times p$, where $h_p=H/p$, $w_p=W/p$, and $N=h_pw_p$.

\noindent\textbf{Dynamic Region Estimation.} Given a pair of multi-modal video sequences $S^1$ and $S^2$, we first compute the temporal mean representation of each modality to characterize temporal reference, as follows:
\begin{equation}
\bar{S}^{m}=\frac{1}{T}\sum_{t=1}^{T} S_t^{m},\qquad m\in\{\mathrm{1},\mathrm{2}\}.
\end{equation}

Based on the predefined patch partition, the dynamic response of the $n$-th patch is calculated by averaging the pixel-wise deviations of the $t$-th frame from the temporal mean:
\begin{equation}
R_{t,n}^{m}=\frac{1}{p^2}\sum_{(u,v)\in\Omega_n}\left| S_t^{m}(u,v)-\bar{S}^{m}(u,v) \right|,
\end{equation}
where $\Omega_n$ denotes the set of spatial locations within the $n$-th patch, and $(u,v)$ denotes a pixel coordinate in $\Omega_n$.

The dynamic responses of the two sequences are combined as:
\begin{equation}
R_{t,n}=R_{t,n}^{\mathrm{1}}+R_{t,n}^{\mathrm{2}}.
\end{equation}

Finally, the dynamic regions $\mathcal{D}_t$ are identified by selecting the Top-$K$ patches of frame t according to their dynamic scores:
\begin{equation}
\mathcal{D}_t=\operatorname{Top\text{-}\mathit{K}}_{n}\left(R_{t,n};K\right).
\end{equation}
where $K=\max\left(1,\operatorname{round}(\rho N)\right)$ denotes the number of selected patches, and $\rho$ is a hyperparameter controlling the proportion of selected dynamic patches.

\noindent\textbf{Spatio-Temporal Interaction.} Given the patch embeddings $Z\in\mathbb{R}^{B\times T\times N\times D}$ output by the VITTT block, STIM directly operates on these token representations, where $D$ denotes the embedding dimension.

As shown in Fig.~\ref{fig2}, STIM consists of two parallel branches: a sparse temporal interaction branch and a dense spatial refinement branch. Guided by $\mathcal{D}$, the sparse branch performs spatio-temporal interaction at the selected patch locations, while the dense branch performs local spatial refinement over the full grid.

Before describing the two branches, we first introduce the TTT-based formulation for inner-model adaptation. Let $f_{\theta}$ denote an arbitrary inner model parameterized by the learnable initial fast weights $\theta$. Given $M$ key--value pairs, the fast weights are adapted once using the dot-product adaptation objective:
\begin{equation}
\mathcal{L}_{\mathrm{adapt}}(\theta;K,V) = -\frac{\gamma}{M} \left\langle f_{\theta}(K),V\right\rangle_F,
\label{eq:adaptloss}
\end{equation}
where $\langle\cdot,\cdot\rangle_F$ denotes the Frobenius inner product. The corresponding one-step fast-weight update is defined as:
\begin{equation}
\theta^{*}=\theta-\nabla_{\theta} \mathcal{L}_{\mathrm{adapt}}(\theta;K,V),
\label{eq:update}
\end{equation}
where $\theta^{*}$ denotes the adapted fast weights obtained after an inner-loop update.

\noindent\emph{Sparse temporal interaction branch.} The selected patch embeddings are first gathered as:
\begin{equation}
Z_s = \operatorname{Gather}\left(Z,\mathcal{D}\right), \qquad Z_s\in\mathbb{R}^{B\times T\times K\times D}.
\end{equation}

To preserve the frame identity of the selected patch embeddings during subsequent spatio-temporal interaction, a sinusoidal temporal positional encoding is added:
\begin{equation}
Z_s^{\mathrm{pos}}=Z_s+P,
\end{equation}
where $P\in\mathbb{R}^{1\times T\times 1\times D}$ denotes the frame-wise temporal positional encoding. The temporal and selected-patch dimensions are then merged to form a token sequence $Z_s^{\mathrm{seq}}\in\mathbb{R}^{B\times TK\times D}$ concatenated along the temporal dimension.

Queries, keys, and values are generated from the spatio-temporal sequence through learnable linear projections:
\begin{equation}
Q_s=Z_s^{\mathrm{seq}}W_{Q_s},\qquad
K_s=Z_s^{\mathrm{seq}}W_{K_s},\qquad
V_s=Z_s^{\mathrm{seq}}W_{V_s},
\end{equation}
where $Q_s,K_s,V_s\in\mathbb{R}^{B\times TK \times D}$.

For the sparse temporal interaction branch, the inner model is implemented as a simplified SwiGLU parameterized by $\theta_s=\{W_1,W_2\}$. For an arbitrary input $U$, it is defined as:
\begin{equation}
f_{\theta_s}(U) = (UW_1)\odot\operatorname{SiLU}(UW_2),
\end{equation}
where $\odot$ denotes element-wise multiplication. Using $K_s$ and $V_s$ as the key--value inputs and substituting $\theta=\theta_s$, $K=K_s$, $V=V_s$, and $M=TK$ into Eq.~\eqref{eq:adaptloss} defines the adaptation objective for this branch. Applying the one-step update in Eq.~\eqref{eq:update} yields the adapted parameter set $\theta_s^{*}=\{W_1^{*},W_2^{*}\}$.

The adapted inner model is then applied to the queries to produce the output of this branch:
\begin{equation}
O_s^{\mathrm{seq}} = f_{\theta_s^{*}}(Q_s) = (Q_sW_1^{*})\odot\operatorname{SiLU}(Q_sW_2^{*}).
\end{equation}

After restoring the temporal and selected-patch dimensions, the interacted embeddings are scattered back to their original spatial locations according to $\mathcal{D}$, where $O_s\in\mathbb{R}^{B\times T\times K\times D}$ denotes the dimension-restored form of $O_s^{\mathrm{seq}}\in\mathbb{R}^{B\times TK\times D}$:
\begin{equation}
\widetilde{Z}_s = \operatorname{Scatter}\left(O_s,\mathcal{D};N\right), \qquad \widetilde{Z}_s\in\mathbb{R}^{B\times T\times N\times D}.
\end{equation}

\noindent\emph{Dense spatial refinement branch.} In parallel with the sparse temporal interaction, the dense branch operates on the full grid and adapts the convolutional inner-model design for local spatial refinement~\cite{VITTT}. The complete patch embeddings are projected into queries, keys, and values, which are subsequently rearranged according to their original patch locations as $Q_l,K_l,V_l\in\mathbb{R}^{BT\times C\times h_p\times w_p}$,
where the batch and temporal dimensions are merged and $C$ denotes the channel dimension.

The convolutional inner model is parameterized by the learnable initial fast weights $\theta_l=\{W_3\}$ and implemented using a $3\times3$ depth-wise convolution:
\begin{equation}
f_{\theta_l}(U) = \operatorname{DWConv}_{3\times3}(U;W_3).
\end{equation}

The full grid key--value features $K_l$ and $V_l$ are used to instantiate the adaptation formulation in Eqs.~\eqref{eq:adaptloss} and~\eqref{eq:update}, with $\theta=\theta_l$, $K=K_l$, $V=V_l$, and $M=N$. This produces the adapted parameter set $\theta_l^{*}=\{W_3^{*}\}$.

The adapted inner model is then applied to the full grid queries to obtain the locally refined features:
\begin{equation}
O_l = f_{\theta_l^{*}}(Q_l) = \operatorname{DWConv}_{3\times3}(Q_l;W_3^{*}),
\end{equation}
where $O_l\in\mathbb{R}^{BT\times C\times h_p\times w_p}$. The resulting feature maps are finally rearranged back into token form, yielding $\widetilde{Z}_l\in\mathbb{R}^{B\times T\times N\times D}$. Finally, the two branch outputs, $\widetilde{Z}_l$ and $\widetilde{Z}_s$, are linearly fused.

\subsection{Loss Function}
\label{sec:loss_function}

\noindent\textbf{Pixel-level Supervision.} To preserve frame-wise intensity and structural information, we adopt two pixel-level constraints: an intensity loss and a gradient loss. Accordingly, the pixel-level objective is defined as:
\begin{equation}
\mathcal{L}_{\mathrm{pixel}} = \lambda_1\mathcal{L}_{\mathrm{int}} +
\lambda_2\mathcal{L}_{\mathrm{grad}},
\label{eq:pixel_loss}
\end{equation}
where $\lambda_{\mathrm{1}}$ and $\lambda_{\mathrm{2}}$ are hyperparameters. Then, the intensity and gradient terms are respectively formulated as:
\begin{equation}
\begin{aligned}
\mathcal{L}_{\mathrm{int}}&=\frac{1}{THW}\sum_{t=1}^{T}\left\|S_t^F-\max\left(S_t^1,S_t^2\right)\right\|_1, \\
\mathcal{L}_{\mathrm{grad}} &=\frac{1}{THW}\sum_{t=1}^{T}\left\|
\left|\nabla S_t^F\right| -\max\left( \left|\nabla S_t^1\right|, \left|\nabla S_t^2\right| \right) \right\|_1.
\end{aligned}
\end{equation}
Here, $\max(\cdot,\cdot)$ denotes element-wise maximum selection, and $\nabla$ denotes the Sobel gradient operator.

\noindent\textbf{Feature-level Supervision.} Although pretrained MAE representations \cite{MaeFuse,MAEDUNet} have been increasingly adopted as visual priors, their patch-level contextual modeling capability remains underexplored for supervising temporal transitions in MMVF. Masked image modeling\cite{MAE} encourages a model to recover masked image regions from their visible context, enabling the learned representations to capture contextual dependencies among image patches. Consequently, features extracted by a pretrained MAE encoder are responsive to variations in structural patterns, object boundaries, and regional content. When applied to video sequences, the temporal variations of these features can effectively characterize the structural- and region-level dynamic evolution of content.

Based on this property, we employ a frozen MAE encoder as a shared representation space for feature-level temporal supervision. Given the fused sequence and the two source sequences, $S^{F}$, $S^{1}$, and $S^{2}$, the MAE feature of each frame is extracted as:
\begin{equation}
E^{m}_t = \Phi(S^{m}_t), \quad m\in\{F,1,2\},
\label{eq:mae_feature}
\end{equation}
where $\Phi$ denotes the frozen MAE encoder, and $E^{F}_t$, $E^{1}_t$, and $E^{2}_t$ represent the features of the fused and the two source sequences at frame $t$, respectively.

To characterize inter-frame feature variations $\dot{E}_t^{m}$, the formulation is given as follows:
\begin{equation}
\dot{E}_t^{m} = E_{t+1}^{m}-E_t^{m}, \qquad 1 \le t < T,\quad m\in\{F,1,2\}.
\label{eq:mae_variations}
\end{equation}

Then, we define a temporal consistency loss to align the temporal transition directions of the fused sequence with those of both source sequences between consecutive frames:
\begin{equation}
\mathcal{L}_{\mathrm{temp}} = \frac{1}{2(T-1)} \sum_{t=1}^{T-1} \sum_{m=1}^{2}
\left[ 1-\cos\left(\dot{E}_t^{F},\dot{E}_t^{m}\right) \right].
\end{equation}
where $\cos(\cdot,\cdot)$ denotes cosine similarity between the corresponding feature variations. 

In addition, we impose a time-reversal consistency constraint on the fused sequence as a self-consistency regularizer to reduce the model's dependence on a specific temporal processing direction. Specifically, the reversed input sequences $S_{\mathrm{re}}^{1}$ and $S_{\mathrm{re}}^{2}$ are fused to obtain $S_{\mathrm{re}}^{F}$. After restoring it to the original temporal order, its feature variations are expected to be consistent with those of the original fusion result $S^{F}$. By applying Eqs.~\eqref{eq:mae_feature} and~\eqref{eq:mae_variations}, we obtain its patch-level feature $E^{\mathrm{re}}$ and the corresponding adjacent-frame temporal feature variations $\dot{E}_t^{\mathrm{re}}$. This constraint suppresses temporal-direction bias and promotes more stable temporal modeling. The time-reversal consistency loss is then defined as:
\begin{equation}
\mathcal{L}_{\mathrm{rev}} = \frac{1}{T-1} \sum_{t=1}^{T-1} \left[ 1 - \cos\left(\dot{E}_t^{F}, \dot{E}_t^{\mathrm{re}} \right) \right].
\end{equation}

Beyond temporal supervision, the frozen MAE features are also used to provide structural and semantic guidance over the intermediate features of the network. Specifically, for each modality, the output features of all three ViTTT blocks are aligned with the final output features of MAE. Let $H_m^l$ denote the output feature of the $l$-th ViTTT block for modality $m\in\{1,2\}$, where $l\in\{1,2,3\}$. Before feature alignment, a modality- and layer-specific projection head maps $H_m^l$ to the MAE feature dimension, yielding $\widehat{H}_m^l$. The distillation loss is defined as:
\begin{equation}
\mathcal{L}_{\mathrm{distill}} = \frac{1}{6T} \sum_{t=1}^{T} \sum_{m=1}^{2}
\sum_{l=1}^{3}\left[1-\cos\left(\widehat{H}_{t,m}^{l},E_t^{m}\right)
\right].
\end{equation}

Therefore, the feature-level objective is defined as:
\begin{equation}
\mathcal{L}_{\mathrm{feat}} = \lambda_3\mathcal{L}_{\mathrm{temp}} + \lambda_4\mathcal{L}_{\mathrm{rev}} + \lambda_5\mathcal{L}_{\mathrm{distill}},
\label{eq:pixel_loss}
\end{equation}
where $\lambda_{\mathrm{3}}$, $\lambda_{\mathrm{4}}$ and $\lambda_{\mathrm{5}}$ are hyperparameters.

Overall, the final training loss of the proposed model is as follows:
\begin{equation}
\mathcal{L}_{\mathrm{total}} = \mathcal{L}_{\mathrm{pixel}} + \mathcal{L}_{\mathrm{feat}} 
\end{equation}

\section{Experiments}
\subsection{Experiment Settings}
\noindent\textbf{Datasets.} The proposed model is trained on three datasets: HDO~\cite{HDO}, M3SVD~\cite{VideoFusion}, and VTUAV~\cite{VTUAV}. For HDO, 22 sequences are used for training, with an average of approximately 240 frames per sequence, while the remaining five sequences are used for testing. For M3SVD, 50 sequences are used for training, with an average of approximately 200 frames per sequence, and eight sequences are selected from the remaining data for testing. For VTUAV, 38 sequences are used for training, with an average of approximately 100 frames per sequence, and eight sequences are selected from the remaining data for testing. To evaluate the generalization ability of the proposed method, eight sequences from the VTMOT dataset~\cite{VF-Bench} are additionally selected for testing. 

\noindent\textbf{Training Details.} All experiments are implemented in PyTorch and conducted on a server equipped with two NVIDIA GeForce RTX 4090 GPUs. The model is trained for a total of 50,000 iterations using a two-stage progressive training strategy. During the first 25,000 iterations, five-frame video clips are randomly cropped to $256\times256$ pixels for training, with a batch size of five. During the remaining 25,000 iterations, the spatial resolution and temporal length are increased to $384\times384$ pixels and seven frames, respectively, while the batch size is reduced to three. AdamW is adopted as the optimizer with an initial learning rate of $1\times10^{-4}$. The patch size is set to $16\times16$ pixels, and the $\rho$ for dynamic-region selection is set to 0.25. The fixed scaling factor $\gamma$ is set to $9^{-\frac{1}{2}}$. The loss weights are empirically set to $\lambda_1=5$, $\lambda_2=10$, $\lambda_3=2$, $\lambda_4=2$, and $\lambda_5=10$.

\noindent\textbf{Comparative Methods.} In this paper, we select seven state-of-the-art fusion methods for comparison, including three video fusion methods (VideoFusion~\cite{VideoFusion}, TemCoCo~\cite{temcoco}, and UniVF~\cite{VF-Bench}) and four image fusion methods (DSPFusion~\cite{DSPFusion}, ISFusion~\cite{ISFusion}, GIFNet~\cite{GIFNet}, and CoCoNet~\cite{CoCoNet}). For fair comparison, all competing methods are evaluated using the parameter settings reported in their original papers.

\noindent\textbf{Evaluation Metrics.} At the image level, we employ $\mathrm{Q}_\mathrm{G}$, $\mathrm{Q}_\mathrm{M}$, FSIM, and VIF to evaluate fusion quality from multiple perspectives~\cite{QGQM,FSIM,VideoFusion}. Specifically, $\mathrm{Q}_\mathrm{G}$ measures the preservation of edge strength and orientation from the source images. $\mathrm{Q}_\mathrm{M}$ evaluates the retention of multi-scale edge and high-frequency information. FSIM assesses the similarity of salient local features and structure. VIF evaluates the fidelity of fusion results in accordance with human visual perception.

For temporal quality assessment, flowD and feaCD are adopted as existing evaluation metrics~\cite{temcoco}. flowD quantifies frame-to-frame temporal alignment errors through optical-flow-based frame prediction. feaCD evaluates the directional consistency of inter-frame variations between the fused and source sequences over the entire frame by mapping each frame into the ResNet-18 output space. In addition, we introduce a new metric (FD3), which quantifies the dynamic direction distance between the inter-frame feature variations of the fused sequence and those of each source sequence within local dynamic regions in the pretrained MAE feature space.

Following the feature-level supervision introduced in Sec.~\ref{sec:loss_function},
a frozen MAE encoder is employed to extract spatially aligned patch features from each frame. Let
\begin{equation}
\mathbf{E}_{t}^{m} = \left[ \mathbf{e}_{t,1}^{m}, \mathbf{e}_{t,2}^{m}, \ldots, \mathbf{e}_{t,N}^{m}
\right]^{\top} \in \mathbb{R}^{N \times D}, \qquad m \in \{F,1,2\},
\label{eq:fd3_patch_feature}
\end{equation}
where $\mathbf{e}_{t,n}^{m} \in \mathbb{R}^{D}$ denotes the MAE patch feature corresponding to the $n$-th spatial region of sequence $m$ at frame $t$. 
For each fixed spatial region, we first compute its temporal mean feature and then derive the feature deviation from this mean:
\begin{equation}
\overline{\mathbf{e}}_{n}^{m} = \frac{1}{T} \sum_{t=1}^{T} \mathbf{e}_{t,n}^{m},
\qquad \Delta \mathbf{e}_{t,n}^{m} = \mathbf{e}_{t,n}^{m} - \overline{\mathbf{e}}_{n}^{m}.
\label{adj-frame var}
\end{equation}
The resulting variations are weak in static regions but pronounced in regions containing moving targets, evolving boundaries, and temporally varying content. Accordingly, FD3 is defined as
\begin{equation}
\operatorname{FD3} = \frac{1}{TN} \sum_{t=1}^{T} \sum_{n=1}^{N} \sum_{m \in \{1,2\}} \left[
1- \cos \left( \Delta \mathbf{e}_{t,n}^{F}, \Delta \mathbf{e}_{t,n}^{m} \right) \right].
\label{eq:fd3}
\end{equation}
A lower FD3 value indicates smaller directional discrepancies between the region-level feature variations of the fused sequence and those of the two source sequences.

\begin{figure*}[t]
   \centering
   \includegraphics[width=1.0\linewidth]{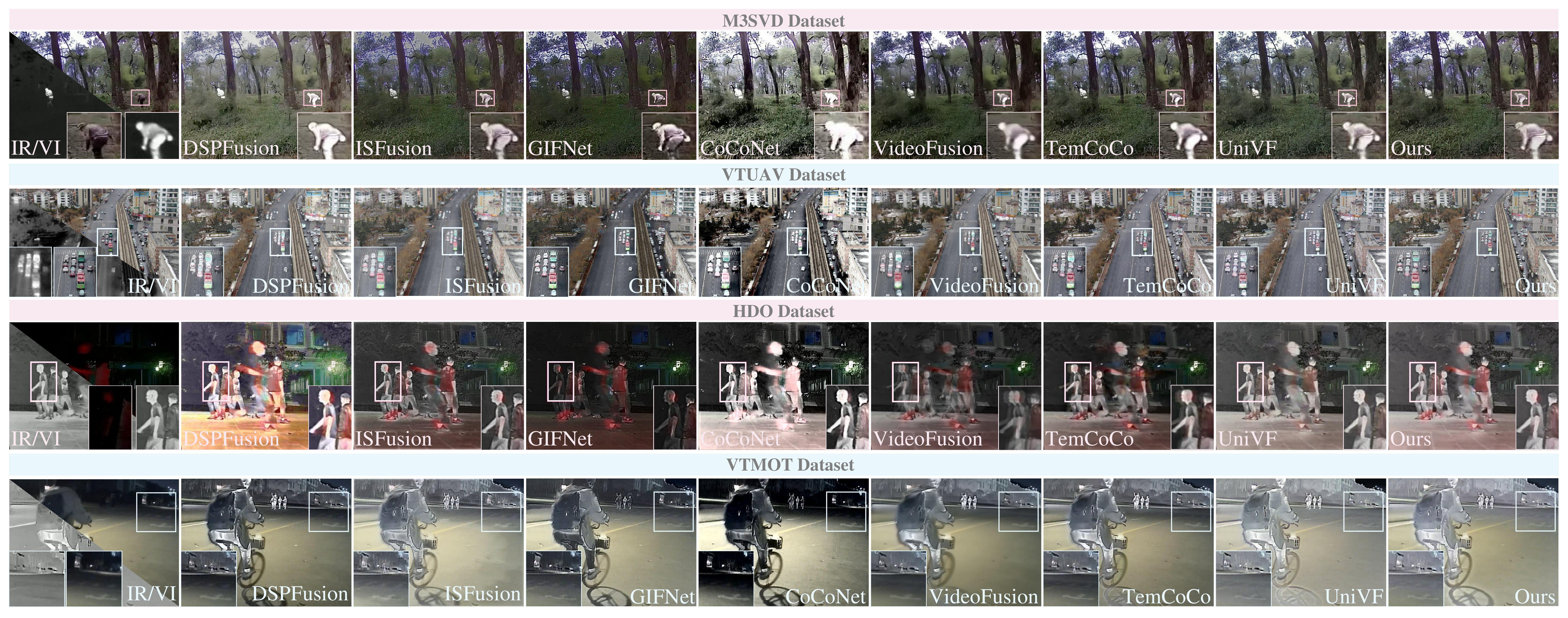}   
   \caption{Qualitative comparison of all methods on the M3SVD, VTUAV, HDO and VTMOT datasets.}
   \label{fig3}   
\end{figure*}

\begin{table*}[t]
\caption{Quantitative comparisons across the M3SVD, VTUAV, HDO, and VTMOT datasets. The best, second-best, and third-best results are denoted by \redmark{bestcolor}, \bluemark{secondcolor}, and \greenmark{thirdcolor}, respectively.} 
\centering
\setlength{\tabcolsep}{3.2pt}
\renewcommand{\arraystretch}{1.10}
\resizebox{\textwidth}{!}{%
\begin{tabular}{c|cl|ccccccc@{\hspace{18pt}}ccccccc}
\hline \multirow{2}{*}{Type} & \multirow{2}{*}{Pub.} & \multirow{2}{*}{Method}
& $\mathrm{Q}_{\mathrm{G}}\uparrow$ & $\mathrm{Q}_{\mathrm{M}}\uparrow$
& $\mathrm{VIF}\uparrow$ & $\mathrm{FSIM}\uparrow$ & $\mathrm{feaCD}\downarrow$
& $\mathrm{flowD}\downarrow$ & $\mathrm{FD3}\downarrow$ & $\mathrm{Q}_{\mathrm{G}}\uparrow$ & $\mathrm{Q}_{\mathrm{M}}\uparrow$
& $\mathrm{VIF}\uparrow$ & $\mathrm{FSIM}\uparrow$ & $\mathrm{feaCD}\downarrow$
& $\mathrm{flowD}\downarrow$ & $\mathrm{FD3}\downarrow$ \\ \cline{4-10}\cline{11-17}
& & & \multicolumn{7}{c}{M3SVD Dataset} & \multicolumn{7}{c}{VTUAV Dataset} \\ \hline

\multirow{4}{*}{Image} & \textit{TIP 26} & DSPFusion & \greencell{0.61} & \greencell{0.58} & \greencell{0.31} & 0.782 & 1.76 & 8.72 & 1.15 & \bluecell{0.57} & \bluecell{0.93} & \greencell{0.34} & 0.793 & \bluecell{1.73} & 12.51 & \greencell{1.09} \\
& \textit{CVPR 26} & ISFusion & 0.51 & 0.35 & 0.25 & \bluecell{0.777} & \bluecell{1.69} & 6.96 & \bluecell{1.12} & 0.48 & 0.74 & 0.33 & 0.808 & 1.77 & 9.35 & \bluecell{1.05} \\
& \textit{CVPR 25} & GIFNet & 0.47 & 0.25 & 0.24 & 0.776 & 1.78 &6.32 & \greencell{1.14} & 0.39 & 0.55 & 0.27 & 0.804 & 1.81 & 10.23 & 1.09 \\
& \textit{IJCV 24} & CoCoNet & 0.35 & 0.19 & 0.18 & 0.697 & 1.76 & 11.15 & 1.30 & 0.33 & 0.41 & 0.21 & 0.739 & 1.83 & 17.47 & 1.18 \\ \hline

\multirow{4}{*}{Video} & \textit{CVPR 26} & VideoFusion & 0.36 & 0.30 & 0.30 & \redcell{0.802} & \greencell{1.70} & \greencell{5.81} & 1.33 & 0.29 & 0.49 & 0.30 & \greencell{0.811} & 1.94 & \bluecell{7.95} & 1.36 \\
& \textit{ICCV 25} & TemCoCo & 0.42 & 0.26 & 0.28 & 0.788 & 1.88 & 8.22 & 1.35 & 0.32 & 0.50 & 0.25 & 0.808 & 1.97 & \redcell{6.70} & 1.37 \\
& \textit{NeurIPS 25} & UniVF & \bluecell{0.63} & \bluecell{0.62} & \bluecell{0.37} & \greencell{0.796} & 1.88 & \redcell{3.46} & 1.13 & \greencell{0.57} & \greencell{0.89} & \bluecell{0.39} & \bluecell{0.813} & \greencell{1.73} & 8.72 & 1.14 \\
& & Ours & \redcell{0.63} & \redcell{0.82} & \redcell{0.39} & \bluecell{0.798} & \redcell{1.66} & \bluecell{5.68} & \redcell{1.06} & \redcell{0.61} & \redcell{1.26} & \redcell{0.41} & \redcell{0.813} & \redcell{1.61} & \greencell{8.07} & \redcell{1.04} \\ \hline

\multicolumn{3}{c|}{} & \multicolumn{7}{c}{HDO Dataset} & \multicolumn{7}{c}{VTMOT Dataset} \\ \hline

\multirow{4}{*}{Image} & \textit{TIP 26} & DSPFusion & \bluecell{0.53} & \bluecell{0.52} & 0.26 & 0.775 & \bluecell{1.76} & 14.47 & \greencell{1.13} & \redcell{0.54} & \bluecell{0.58} & \greencell{0.33} & 0.804 & \bluecell{1.70} & 19.66 & \greencell{1.05} \\
& \textit{CVPR 26} & ISFusion & \greencell{0.46} & 0.36 & 0.24 & 0.772 & \greencell{1.77} & 11.38 & \bluecell{1.03} & 0.42 & 0.37 & 0.27 & 0.774 & 1.77 & 15.00 & \bluecell{1.03} \\
& \textit{CVPR 25} & GIFNet & 0.32 & 0.23 & 0.19 & 0.775 & 1.83 & 11.80 & 1.25 & 0.31 & 0.23 & 0.21 & 0.783 & 1.85 & 15.84 & 1.13 \\
& \textit{IJCV 24} & CoCoNet & 0.28 & 0.20 & 0.17 & 0.722 & 1.84 & 16.58 & 1.29 & 0.23 & 0.19 & 0.15 & 0.678 & 1.88 & 23.60 & 1.17 \\ \hline

\multirow{4}{*}{Video} & \textit{CVPR 26} & VideoFusion & 0.30 & 0.30 & \greencell{0.26} & \greencell{0.803} & 1.91 & \redcell{6.41} & 1.32 & 0.27 & 0.27 & 0.29 & \greencell{0.815} & 1.89 & \redcell{9.82} & 1.26 \\
& \textit{ICCV 25} & TemCoCo & 0.35 & 0.28 & 0.23 & 0.801 & 1.87 & \bluecell{9.66} & 1.30 & 0.32 & 0.27 & 0.24 & 0.808 & 1.89 & \bluecell{13.16} & 1.20 \\
& \textit{NeurIPS 25} & UniVF & 0.44 & \greencell{0.41} & \bluecell{0.27} & \bluecell{0.803} & 1.78 & 11.02 & 1.24 & \greencell{0.43} & \greencell{0.49} & \bluecell{0.34} & \bluecell{0.816} & \greencell{1.74} & 14.24 & 1.10 \\
& & Ours & \redcell{0.55} & \redcell{0.65} & \redcell{0.31} & \redcell{0.809} & \redcell{1.71} & \greencell{10.74} & \redcell{1.03} & \bluecell{0.52} & \redcell{0.60} & \redcell{0.37} & \redcell{0.816} & \redcell{1.67} & \greencell{14.15} & \redcell{1.01} \\ \hline
\end{tabular}%
}
\label{tab1} 
\end{table*}

\subsection{Multi-modal Video Fusion Experiment}
\noindent\textbf{Qualitative Comparison.} As shown in Fig.~\ref{fig3}, we present four representative scenes from different datasets. In the first case, different fusion methods are compared in terms of their ability to depict camouflaged persons in a jungle environment. The results of CoCoNet, VideoFusion, and TemCoCo tend to excessively preserve the salient information from infrared images, leading to the loss of visible details. In contrast, the proposed method achieves a better balance between cross-modal information. It not only highlights camouflaged persons in jungle scenes with dense vegetation and complex background textures, but also preserves fine visible details, such as clothing wrinkles and hats. Similarly, in the second case, under sufficient illumination conditions, DSPFusion, CoCoNet, and VideoFusion fail to achieve desirable information balance, resulting in degraded vehicle details. Our method moderately enhances the saliency of vehicles on the road while preserving their original structural details. In the third case, obvious artifacts can be observed in the results of VideoFusion and TemCoCo. These artifacts mainly arise from abnormal residual contents propagated from adjacent frames, indicating the issue of cross-frame information leakage. The proposed method effectively integrates and decouples temporal information, and achieves better contrast enhancement and structural information preservation under insufficient illumination. In the last case, the enlarged regions show that, compared with other methods, the proposed method exhibits stronger detail preservation capability for weak-texture regions, such as zebra crossings on the road and traffic signs. Overall, from the perspective of visual evaluation, our method achieves better performance in terms of contrast, structural information preservation, and cross-modal information balance.

\noindent\textbf{Quantitative Comparison.} Table~\ref{tab1} reports the quantitative comparison between the proposed method and seven state-of-the-art methods on multiple objective metrics. Overall, the proposed method achieves clear advantages on most metrics. For image quality evaluation, all results of our method rank first except for FSIM on the M3SVD dataset and $\mathrm{Q_G}$ on the VTMOT dataset, where our method ranks second. The higher $\mathrm{Q_G}$ and $\mathrm{Q_M}$ values indicate that the proposed method preserves richer textures and detailed information. The advantage in FSIM demonstrates that the fused results maintain strong feature-level consistency with the source sequences. Moreover, the best VIF values suggest that the fused results provide better visual fidelity in terms of perceptually relevant information, such as contrast, color and appearance details.

For temporal evaluation, our method ranks first on all datasets in terms of feaCD and FD3, indicating that it can better maintain source-consistent temporal variation directions both over the global representation and within the local dynamic regions. For flowD, although the proposed method does not achieve the best results on all datasets, it consistently ranks among the top three. Specifically, it obtains the second-best result on the M3SVD dataset, only behind UniVF, and ranks third on the other three datasets, following VideoFusion and TemCoCo. These results demonstrate the strong competitiveness of our method in matching the motion continuity of the source sequences. VideoFusion and TemCoCo achieve lower flowD values, which may partly benefit from over-smoothed fused results. In Sec.~\ref{FD3}, we further analyze the sensitivity of different temporal metrics to temporal smoothing to provide a more comprehensive discussion.

\begin{figure}[t]
   \centering
   \includegraphics[width=1\columnwidth]{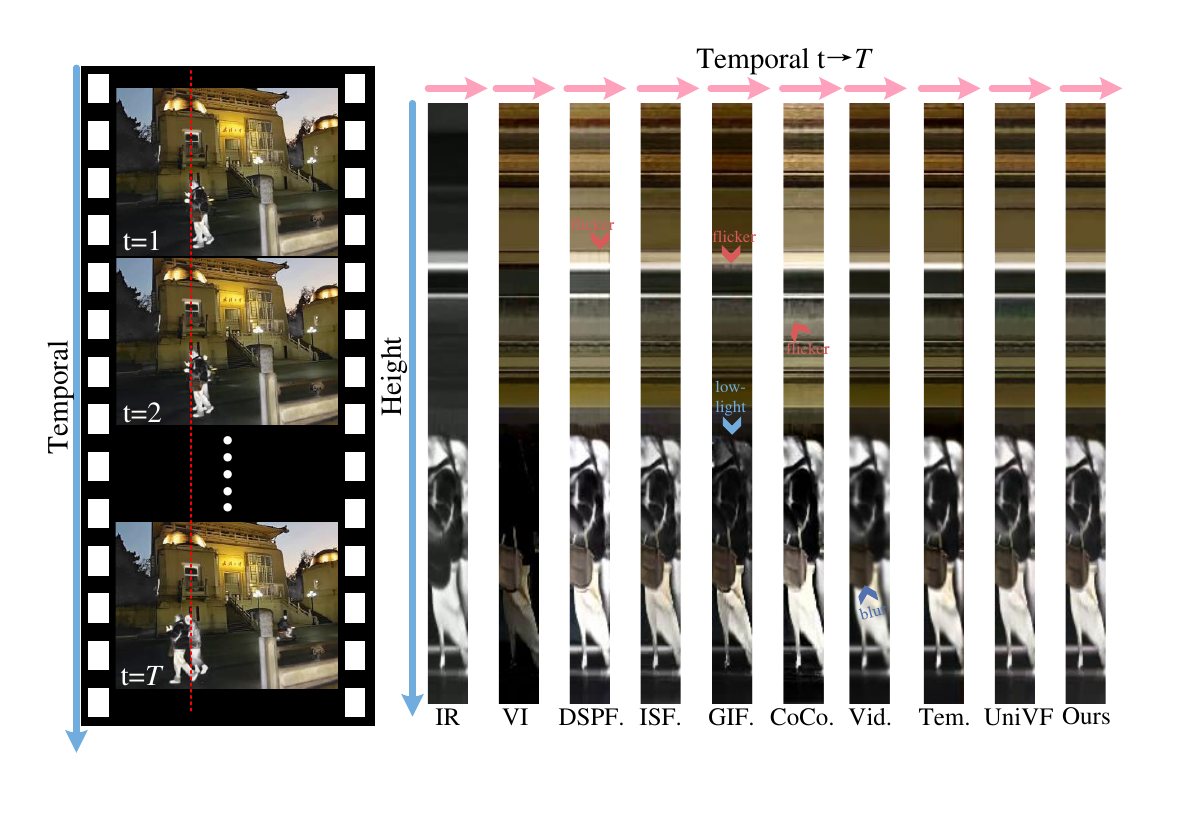}
   \vspace{-30pt}
   \caption{Qualitative comparisons of the temporal consistency for source and fused videos in Scene '0111\_1803' from the M3SVD dataset. We visualize the pixels of the selected column (the red line).}
   \label{fig4}
\end{figure}

\begin{figure}[t]
   \centering
   \includegraphics[width=1\columnwidth]{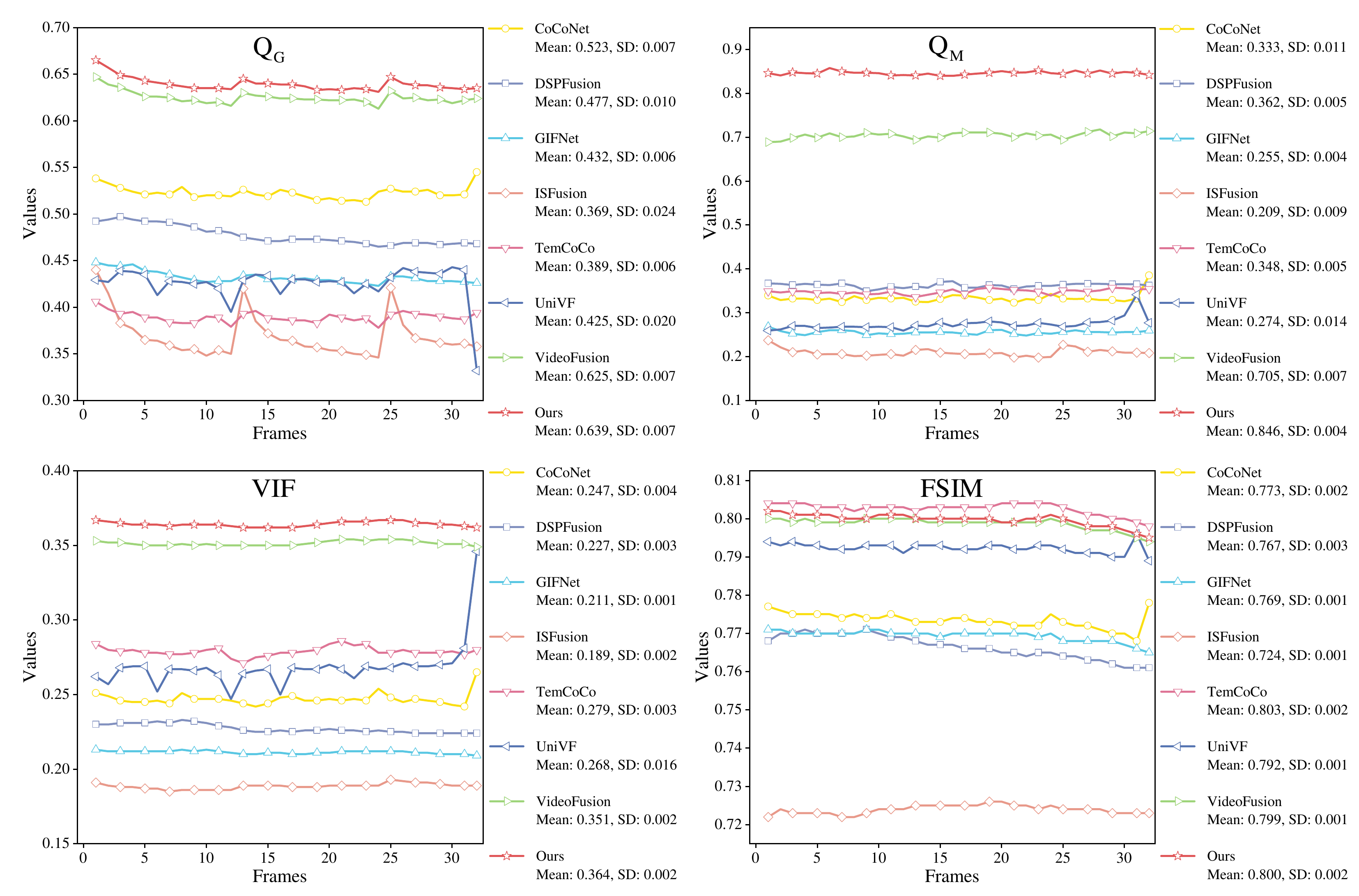}
   \vspace{-15pt}
   \caption{Temporal curves of different metrics over consecutive frames of the sequence from Scene '0111\_1803' of the M3SVD dataset. Mean denotes average value and SD represents standard deviation.}
   \label{fig5}
\end{figure}

\noindent\textbf{Temporal Comparison.} As illustrated in Fig.~\ref{fig4}, the pixels along the fixed column indicated by the red line are extracted from each fused video sequence and temporally concatenated to provide an intuitive visualization of the temporal variations produced by different fusion methods. It can be observed that several image fusion methods suffer from pronounced temporal flickering, particularly DSPFusion and GIFNet. Although VideoFusion achieves favorable performance in terms of temporal metrics, its fused results tend to be overly smooth, leading to degraded edge sharpness and reduced clarity of local details. Furthermore, Fig.~\ref{fig5} presents the frame-wise scores of different evaluation metrics over consecutive frames. The metric curves of several methods exhibit noticeable fluctuations, with UniVF and ISFusion showing particularly unstable behavior. Their relatively large standard deviations further confirm such temporal variations from a statistical perspective. Overall, the proposed method demonstrates consistently stable performance in both subjective visual comparisons and objective quantitative evaluations.

\begin{figure}[t]
   \centering
   \includegraphics[width=0.9\columnwidth]{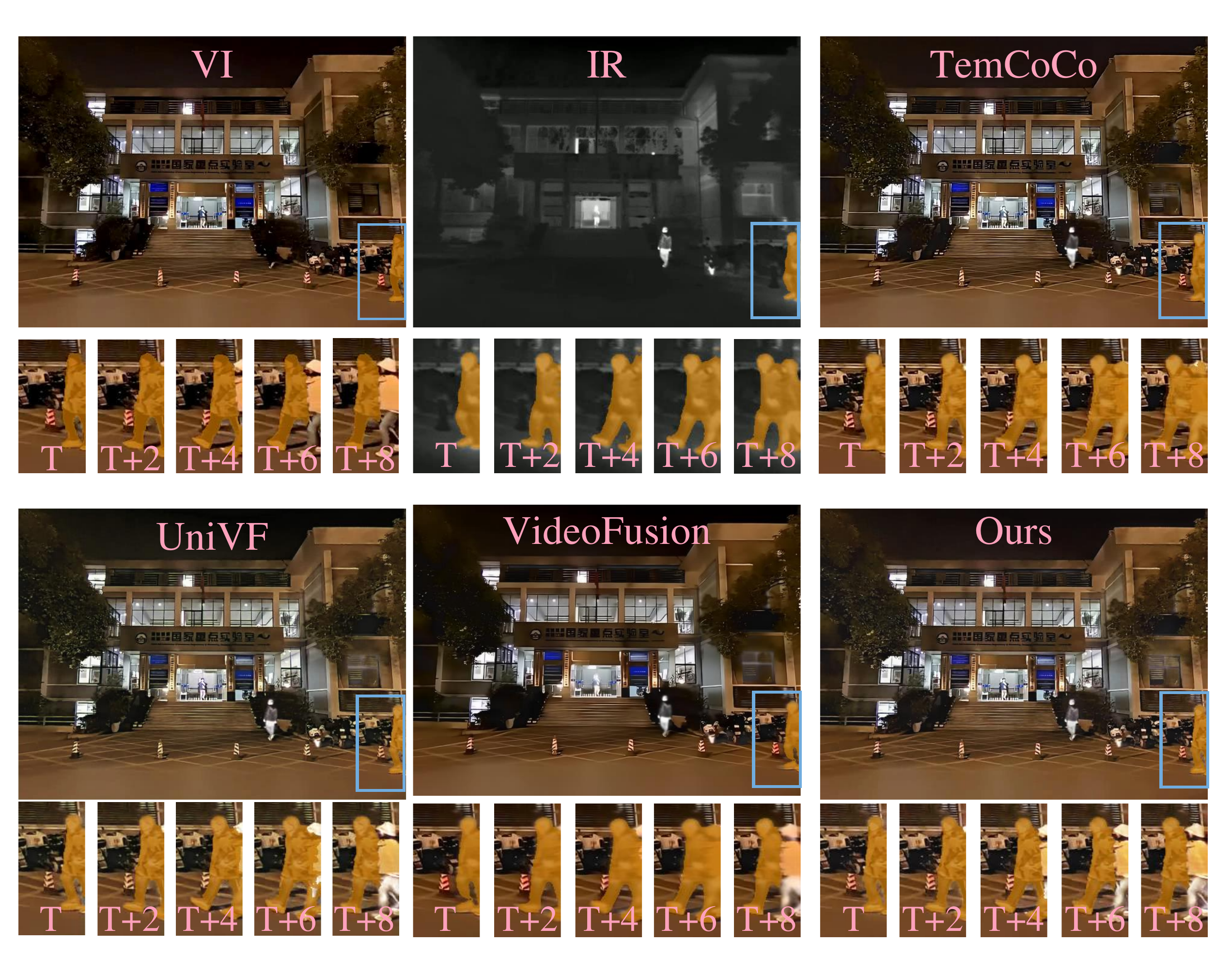}
   \vspace{-13pt}
   \caption{Comparison of segmentation results produced by different video fusion methods on scene ‘0118\_1913’ of the M3SVD dataset. The bottom row of each image presents cropped views of the blue-boxed region at two-frame intervals from $T$ to $T+8$. The segmentation prompt used for this scene is 'the person on the right side wearing a hoodie.'}
   \label{fig5.5}
\end{figure}

\noindent\textbf{Downstream Task Comparison.} To further validate the advantage of the proposed method in temporal stability, we employ SAMWISE~\cite{SAMWISE} as a unified segmentation model to perform video segmentation on the sequences generated by different video fusion methods. As shown in Fig.~\ref{fig5.5}, at time $T+6$, all comparison methods exhibit varying degrees of segmentation errors, whereas our method still maintains relatively accurate target segmentation. This demonstrates that our method preserves smoother inter-frame transitions in the fused sequence, thereby providing more stable visual information for downstream segmentation tasks.

\begin{figure}[t]
   \centering
   \includegraphics[width=0.8\columnwidth]{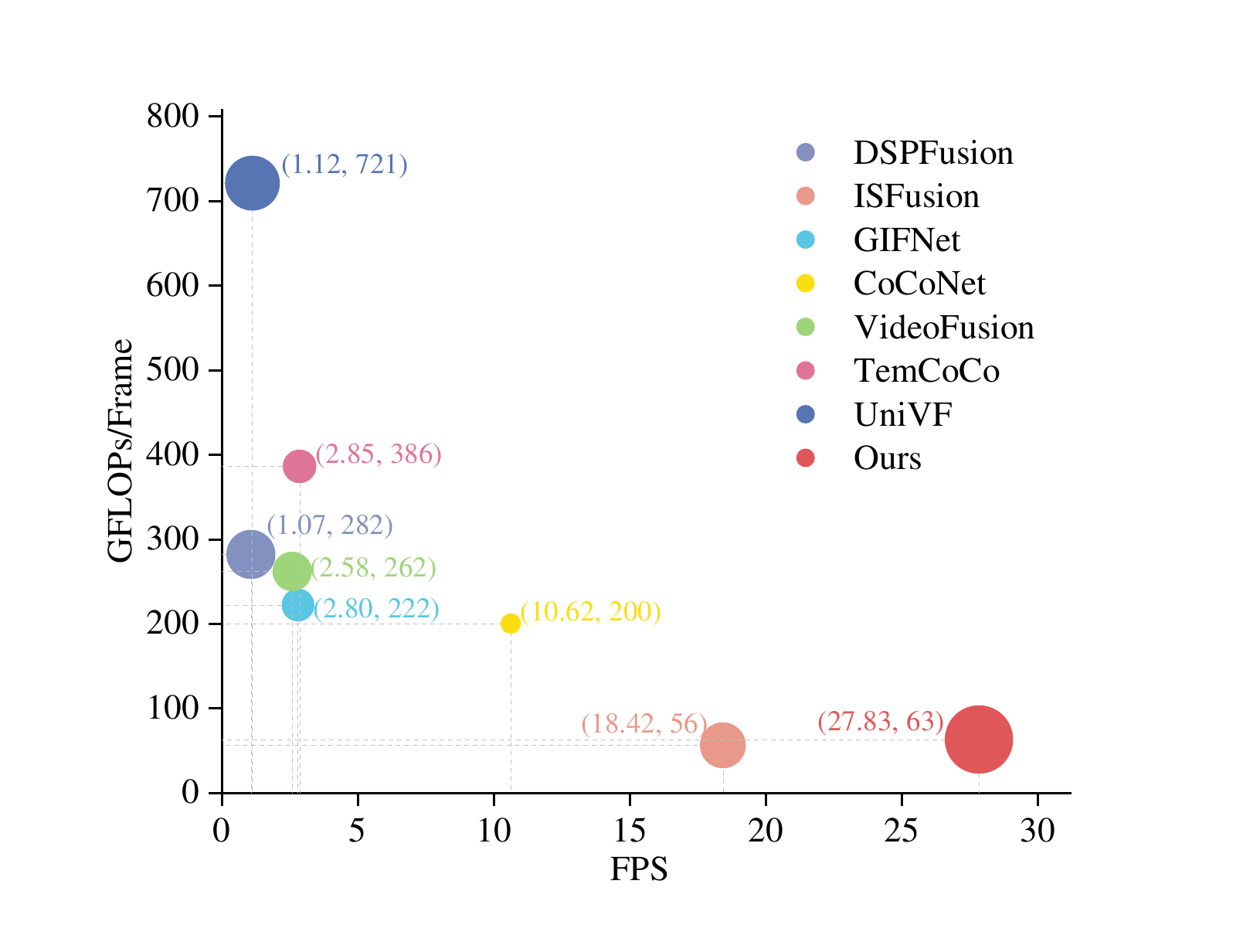}
   \vspace{-13pt}
   \caption{Bubble chart comparing the computational efficiency and fusion performance of different methods. The bubble size is determined by the average rank of each method over all datasets and evaluation metrics, where a larger bubble denotes better overall performance.}
   \label{fig6}
\end{figure}

\noindent\textbf{Efficiency Comparison.} To evaluate the practical inference efficiency of different methods under resource-constrained conditions, we further conduct runtime experiments on an NVIDIA Jetson AGX Orin 64GB Developer Kit. For a fair comparison, all methods are evaluated using FP32 inference under the same software environment and timing protocol, using FP32 tensors with TF32 acceleration enabled on supported NVIDIA GPUs, at an input resolution of 640×480. As shown in Fig.~\ref{fig6}, although UniVF achieves competitive fusion performance, it does so at the expense of substantially reduced computational efficiency. In contrast, our method achieves the highest inference speed of 27.83 FPS while requiring only 63 GFLOPs/frame, representing the second-lowest computational cost among all compared methods. These results demonstrate a superior fusion performance–speed trade-off for real-time inference on the edge platform.

\begin{figure}[t]
   \centering
   \includegraphics[width=1\columnwidth]{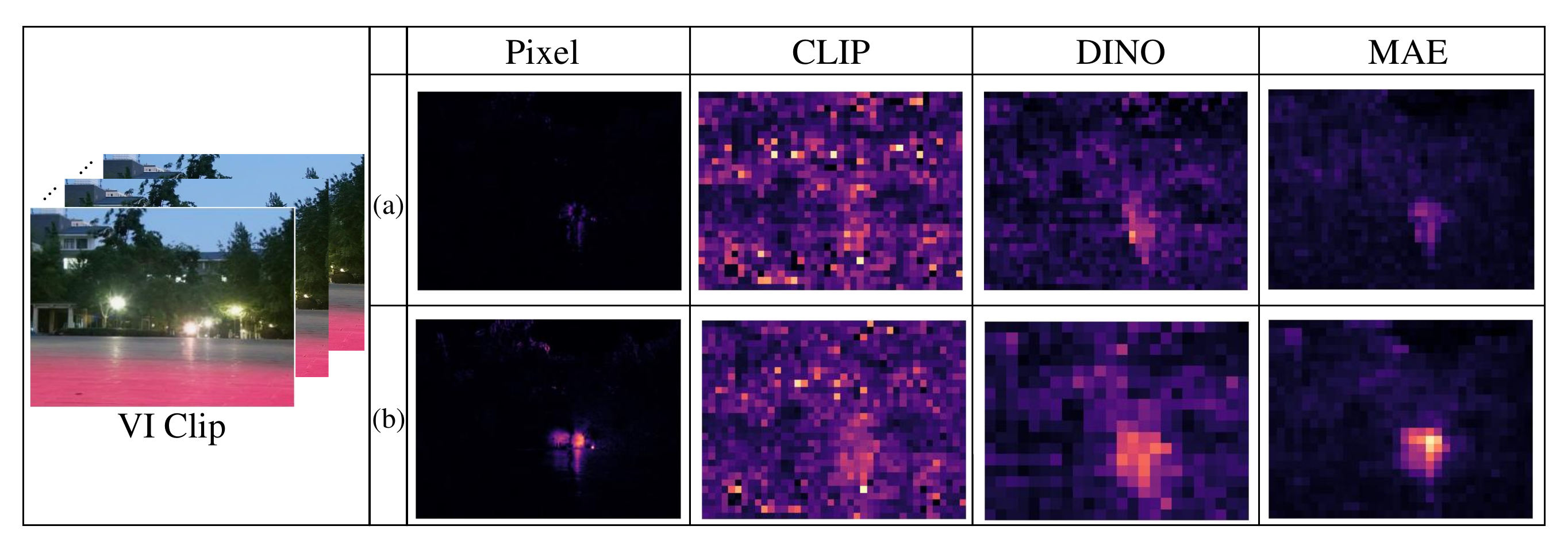}
   \vspace{-15pt}
   \caption{Visualization of variation responses in the pixel space and different pretrained feature spaces for Scene '002' from the HDO dataset. (a) Adjacent-frame variation. (b)  Variation from the temporal mean.}
   \label{fig7.5}
\end{figure}

\begin{table}[t]
\centering
\caption{IoU of the Top-$K$ high-response regions selected from the
pixel-level variation map and different pretrained feature maps.
The best result in each row is highlighted in bold.}
\label{tab2}
\begin{tabular}{c|ccc}
\hline
       & CLIP & DINO & MAE \\ \hline
$\rho=0.10$  & 0.21 & 0.40 & \textbf{0.58} \\
$\rho=0.20$  & 0.30 & 0.46 & \textbf{0.60} \\
$\rho=0.25$  & 0.33 & 0.48 & \textbf{0.60} \\
$\rho=0.30$  & 0.36 & 0.49 & \textbf{0.60} \\
$\rho=0.40$  & 0.42 & 0.52 & \textbf{0.62} \\ \hline
\end{tabular}
\end{table}

\subsection{Analysis of the MAE Feature Space} 

To investigate whether the MAE feature space can more effectively characterize structural- and region-level variations when applied to video sequences, we conduct comprehensive analyses by comparing MAE with other pretrained models.
For the qualitative analysis, as illustrated in Fig.~\ref{fig7.5}, MAE exhibits the highest spatial consistency with the pixel-level responses among the compared pretrained models under both variation formulations, particularly in the high-response regions. For the quantitative analysis, we further evaluate the spatial consistency between the high-response regions obtained from different pretrained feature spaces and the corresponding pixel-level responses using IoU. As shown in Table~\ref{tab2}, MAE consistently achieves the highest IoU across different $\rho$ settings, substantially outperforming the other pretrained models. These qualitative and quantitative results jointly demonstrate that MAE provides a stronger capability for characterizing structural- and region-level variations in video sequences.

\begin{table}[t]
\centering
\caption{Quantitative comparison with different $\rho$ on HDO dataset. The best and second-best results are denoted by \redmark{bestcolor} and \bluemark{secondcolor}, respectively.}
\setlength{\tabcolsep}{2.7pt}
\begin{tabular}{c|ccccccc}
\hline
        & QG$\uparrow$             & QM$\uparrow$             & VIF$\uparrow$            & FSIM$\uparrow$           & feaCD$\downarrow$          & flowD$\downarrow$           & FD3$\downarrow$           \\ \hline
$\rho=0.10$  & \bluecell{0.553}          & \bluecell{0.646}          & \redcell{0.310}          & 0.808          & 1.715          & 10.795          & 1.041       \\

$\rho=0.25$ & \redcell{0.554} & \redcell{0.649}          & \bluecell{0.308}          & \redcell{0.809} & \redcell{1.712}          & \bluecell{10.738}          & \redcell{1.034}          \\ 

$\rho=0.50$  & 0.549          & 0.640          & 0.307          & \bluecell{0.809}          & \bluecell{1.713} & \redcell{10.650} & \bluecell{1.037}                   \\

$\rho=1.00$  & 0.544          & 0.615 & 0.303 & 0.807          & 1.721          & 10.751          & 1.038                   \\
\hline
\end{tabular}
\label{tab2.5}
\end{table}
\subsection{Analysis of the Hyperparameter $\rho$} 

As shown in Table~\ref{tab2.5}, compared with full-region temporal modeling without dynamic region guidance ($\rho=1.00$), the guided variants achieve better performance on the objective evaluation. This indicates that dynamic region guidance enables STIM to focus temporal interactions on tokens containing more informative variations, thereby improving the effectiveness of temporal modeling. Among the different dynamic-region ratios, our method ($\rho=0.25$) achieves the best overall performance.

\begin{table}[t]
\caption{Sensitivity comparison of temporal metrics under different degradation types. Each degradation type is evaluated under progressively more challenging settings. Values in parentheses denote relative changes with respect to the clean baseline. The baseline values are obtained from the proposed method on the HDO dataset without degradation. The most desirable response in each row, jointly considering sensitivity, correct response direction, and monotonicity, is highlighted in bold.}
\centering
\setlength{\tabcolsep}{3pt}
\begin{tabular}{c|c|ccc}
\hline
Perturbation               & Setting  & feaCD$\downarrow$ & flowD$\downarrow$  & FD3$\downarrow$   \\ \hline
Clean                      & Baseline & 1.71 & 10.74 & 1.03 \\ \hline
\multirow{3}{*}{Misalign}  & t=1      & 2.04 (+19.0\%) & 10.85 (+1.1\%)  & \textbf{1.36 (+31.1\%)} \\
                           & t=2      & 2.02 (+18.0\%)  & 11.19 (+4.2\%)  & \textbf{1.50 (+44.9\%)}  \\
                           & t=3      & 1.99 (+16.0\%)  & 11.39 (+6.0\%)  & \textbf{1.59 (+53.6\%)}  \\ \hline
\multirow{3}{*}{Local Lag} & area=0.6 & 1.97 (+14.9\%)  & 11.22 (+4.5\%)  & \textbf{1.38 (+34.0\%)}  \\
                           & area=0.5 & 1.96 (+14.5\%)  & 11.16 (+3.9\%)  & \textbf{1.33 (+29.1\%)}  \\
                           & area=0.4 & 1.96 (+14.2\%)  & 11.08 (+3.2\%)  & \textbf{1.28 (+24.2\%)}  \\ \hline
\multirow{3}{*}{Smoothing} & windows=3 & 1.92 (+12.1\%)  & 6.39 (-40.5\%)   & \textbf{1.16 (+12.5\%) } \\
                           & windows=5 & 1.97 (+15.1\%)  & 4.77 (-55.6\%)   & \textbf{1.24 (+19.9\%)}  \\
                           & windows=7 & 1.97 (+15.2\%)  & 3.90 (-63.7\%)   & \textbf{1.30 (+26.1\%)}  \\ \hline
\end{tabular}
\label{tab3}
\end{table}

\begin{table}[t]
\caption{Sensitivity comparison of FD3 using CLIP~\cite{CLIP}, DINO~\cite{DINOv2}, and MAE as pretrained feature extractors under local content lag perturbations. Percentages in parentheses denote relative changes from the clean baseline. The baseline values are obtained by evaluating each metric on the unperturbed fusion results of the proposed method on the HDO dataset. The largest relative increase from the corresponding clean baseline is highlighted in bold.}
\centering
\begin{tabular}{c|cccc}
\hline
     & Baseline & area=0.60 & area=0.50 & area=0.40 \\ \hline
CLIP & 1.72    &  1.84 (+7\%)     & 1.83 (+6\%)     & 1.81 (+6\%)     \\
DINO & 1.35    &  1.60 (+19\%)    & 1.57 (+17\%)      & 1.54 (+14\%)     \\
MAE  & 1.03    &  \textbf{1.38 (+34\%)}    & \textbf{1.33 (+29\%)}     & \textbf{1.28 (+24\%)}    \\ \hline
\end{tabular}
\label{tab4}
\end{table}

\subsection{Analysis of the FD3 Metric} 
\label{FD3}
As shown in Table~\ref{tab3}, we construct a stress benchmark for the temporal metrics comprising three types of perturbations: temporal misalignment, local content lag, and temporal smoothing. By progressively introducing temporal degradations of different types and severity levels, the benchmark systematically evaluates the degradation-awareness capability of temporal quality metrics in terms of sensitivity, response-direction consistency, and monotonicity. On this basis, FD3 is compared with two existing temporal metrics, flowD and feaCD, to assess their ability to perceive different forms of temporal degradation. The detailed generation procedures for these perturbations are provided in the supplementary material.

Under temporal misalignment, flowD changes by only 1.1\%–6.0\% relative to the unperturbed baseline, indicating limited sensitivity to this type of degradation. feaCD exhibits a more pronounced deviation from the baseline, suggesting that it can detect temporal misalignment. However, as the temporal offset increases from one to three frames, its relative increase unexpectedly decreases from 19.0\% to 16.0\%. The non-monotonic response is inconsistent with the increasing degradation severity. In contrast, FD3 not only exhibits a substantially larger deviation from the baseline, but also shows a continuously increasing response from 31.1\% to 53.6\% as the misalignment becomes more severe. This demonstrates its stronger capability to distinguish different levels of temporal misalignment.

Under local content lag, both flowD and feaCD can, to some extent, discriminate the perturbed sequences from the original baseline. Nevertheless, their responses vary only marginally with the area of the affected region, indicating limited capability to differentiate the spatial extent of local degradation. By comparison, FD3 exhibits a clearer severity-dependent response. As the affected-region ratio increases from 0.4 to 0.6, its relative increase rises from 24.2\% to 34.0\%, demonstrating greater sensitivity and more desirable monotonic behavior.

Under temporal smoothing, flowD continuously decreases as the smoothing strength increases. Since a lower flowD value is interpreted as better temporal quality, this behavior implies that more severe oversmoothing is incorrectly assigned a better evaluation score. This observation suggests that flowD tends to favor smoother inter-frame transitions, but such a preference fails to properly reflect the temporal quality degradation caused by excessive smoothing. Although the response direction of feaCD is generally consistent with expectation, its relative increase grows only from 12.1\% to 15.2\% and gradually saturates under stronger smoothing levels, thereby limiting its ability to distinguish different degrees of temporal smoothing. In contrast, the relative increase of FD3 steadily rises from 12.5\% to 26.1\%, allowing it to continuously capture the temporal quality degradation induced by increasingly severe smoothing.

Across all three perturbations, feaCD exhibits relatively weak responses to changes in degradation severity, which may be attributed to its global information aggregation mechanism. In particular, when static backgrounds occupy a large proportion of the test sequence, temporal degradations occurring in local dynamic regions can be diluted by the dominant stable background information, thereby weakening the metric response to the severity of localized temporal variations.

On the other hand, Table~\ref{tab4} further investigates the influence of pretrained feature spaces on the local degradation sensitivity of FD3. Under local content lag, CLIP exhibits only weak responses to changes in the affected-region ratio, whereas DINO produces more noticeable variations. MAE consistently yields the largest relative changes across all settings. Meanwhile, as the affected-region ratio decreases, the relative change of MAE-based FD3 progressively declines from 34\% to 24\%, forming a clearer monotonic hierarchy. These results suggest that MAE is the most suitable pretrained feature extractor for FD3, enabling more sensitive and discriminative responses to local temporal variations across different spatial extents.

Overall, FD3 is not intended as a simple replacement for existing temporal metrics. Instead, it provides a more interpretable temporal diagnostic perspective that is better aligned with the requirements of MMVF. The stress-benchmark results demonstrate that FD3 consistently exhibits higher sensitivity, more reliable response directions, and clearer monotonic trends across all degradation types, validating its effectiveness in quantifying localized discrepancies in dynamic variation directions between the fused and source sequences and thereby complementing existing temporal metrics.

\begin{table}[t]
\caption{Ablation study of architectural variants, pretrained feature spaces, and loss terms on the HDO dataset. The best results are highlighted in bold.}
\centering
\setlength{\tabcolsep}{2.3pt}
\begin{tabular}{c|ccccccc}
\hline
& $\mathrm{Q}_{\mathrm{G}}$ & $\mathrm{Q}_{\mathrm{M}}$
& $\mathrm{VIF}$ & $\mathrm{FSIM}$ & $\mathrm{feaCD}$
& $\mathrm{flowD}$ & $\mathrm{FD3}$ \\ \hline
w ViT           & 0.551          & 0.635          & 0.306          & 0.808          & 1.713          & 10.771          & 1.037          \\
w/o STIM       & 0.543          & 0.633          & 0.305          & 0.809          & 1.717          & 10.537          & 1.052          \\
w CLIP          & 0.522          & 0.621          & 0.309          & 0.809 & 1.728          & \textbf{10.427} & 1.052          \\
w DINO          & 0.539          & 0.643          & \textbf{0.321}          & \textbf{0.812}          & 1.713          & 10.476          & 1.051          \\
w/o $\mathcal{L}_{\mathrm{rev}}$       & 0.553          & 0.641          & 0.308 & 0.809          & 1.720          & 10.764          & 1.035          \\
w/o $\mathcal{L}_{\mathrm{temp}}$\&$\mathcal{L}_{\mathrm{rev}}$ & 0.536          & 0.628          & 0.302          & 0.806          & 1.722          & 10.984          & 1.055          \\
Ours            & \textbf{0.554} & \textbf{0.649} & 0.308          & 0.809          & \textbf{1.712} & 10.738          & \textbf{1.034} \\ \hline
\end{tabular}
\label{tab5}
\end{table}

\begin{figure}[t]
   \centering
   \includegraphics[width=1.0\columnwidth]{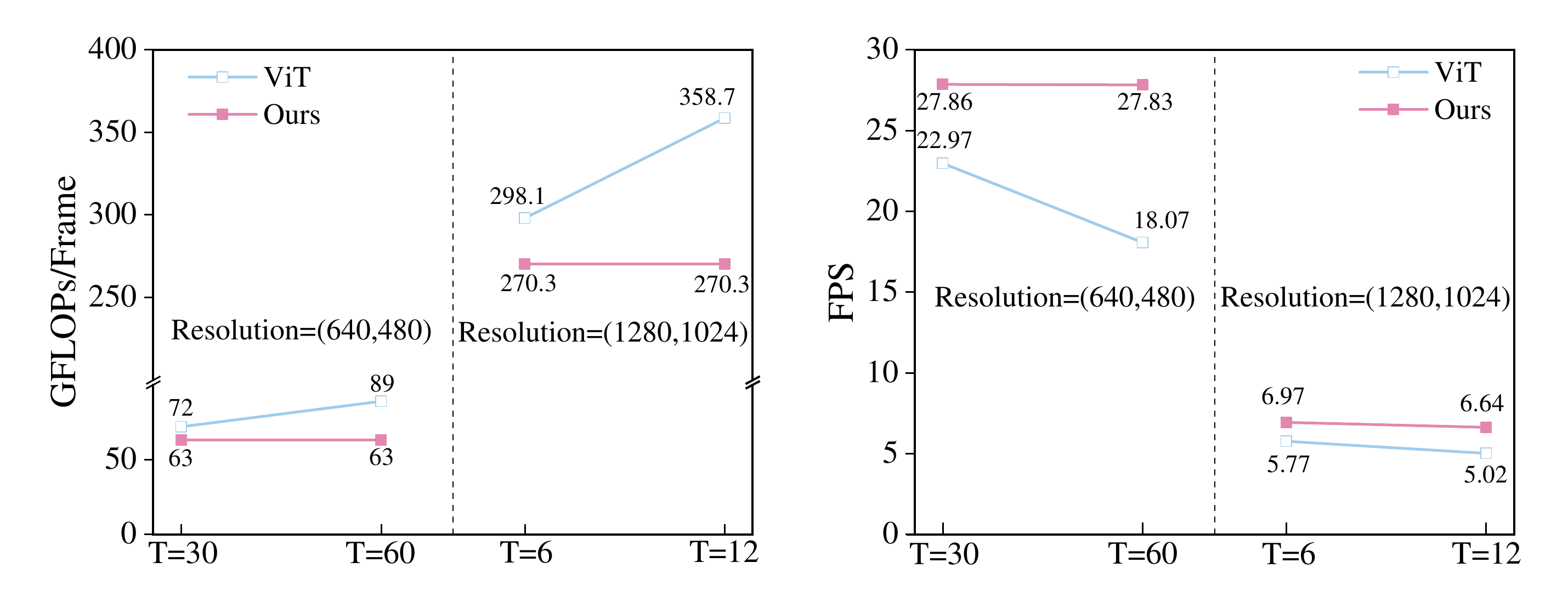}
   \vspace{-10pt}
   \caption{Computational efficiency of ViT and Ours with respect to the number of input frames at two spatial resolutions. For each line chart, the left and right groups correspond to resolutions of $640\times480$ and $1280\times1024$, respectively.}
   \label{fig7} 
\end{figure}

\begin{figure}[t]
   \centering
   \includegraphics[width=0.85\columnwidth]{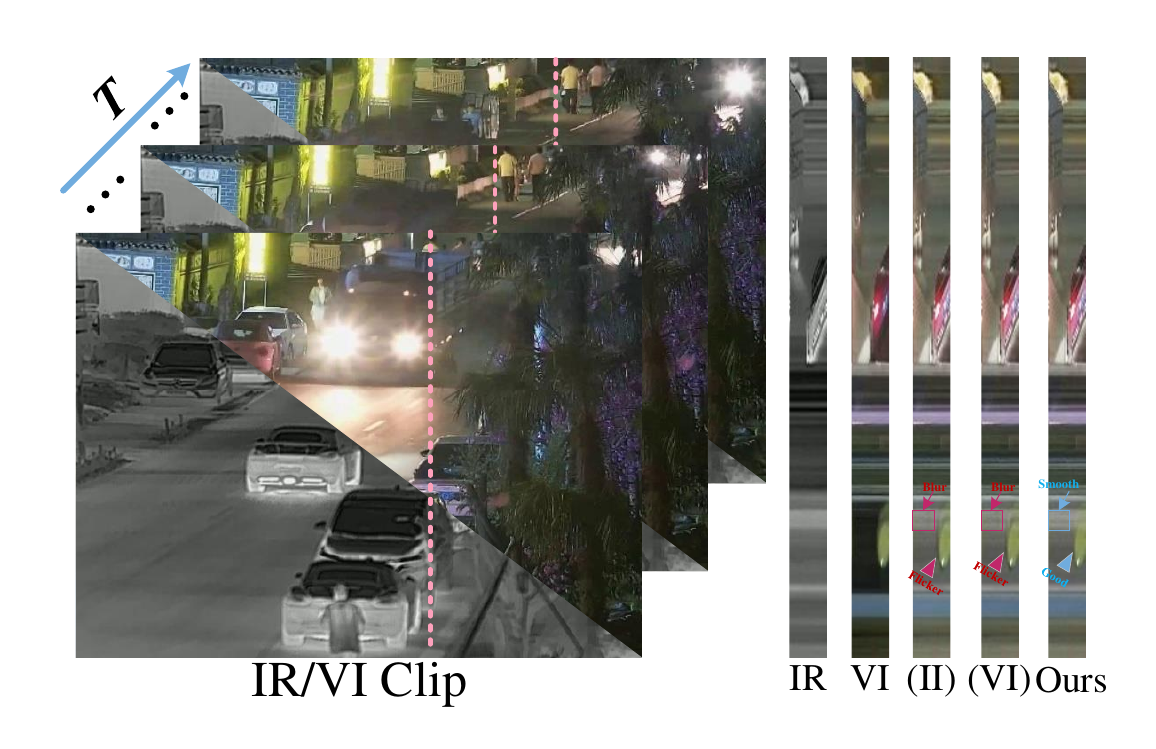}
   \vspace{-10pt}
   \caption{Visual comparison of temporal consistency between two ablation variants: (II) w/o STIM and (VI) w/o $\mathcal{L}_{\mathrm{temp}}$\&$\mathcal{L}_{\mathrm{rev}}$.}
   \label{fig8}
\end{figure}

\subsection{Ablation Experiment} 
To validate the effectiveness of the key components in the proposed method, we conduct six ablation experiments: (I) \textbf{w ViT}: retaining the dynamic region-guided mechanism while replacing the proposed STIM with a conventional ViT; (II) \textbf{w/o STIM}: directly removing STIM; (III) \textbf{w CLIP}: replacing MAE with CLIP as the feature extractor for the temporal loss; (IV) \textbf{w DINO}: replacing MAE with DINO as the feature extractor for the temporal loss; (V) \textbf{w/o $\mathcal{L}_{\mathrm{rev}}$}: removing the $\mathcal{L}_{\mathrm{rev}}$ loss; and (VI) \textbf{w/o $\mathcal{L}_{\mathrm{rev}}$ \& $\mathcal{L}_{\mathrm{temp}}$}: simultaneously removing both $\mathcal{L}_{\mathrm{rev}}$ and $\mathcal{L}_{\mathrm{temp}}$.

As shown in Table~\ref{tab5}, replacing STIM with ViT results in a slight degradation in the overall quantitative performance. More importantly, as illustrated in Fig.~\ref{fig7}, the computational cost of ViT increases rapidly as the number of input frames or the spatial resolution grows, accompanied by a substantial reduction in inference speed. In contrast, STIM employs a linear-complexity sequence modeling mechanism and therefore maintains relatively stable computational efficiency across different input scales.

For experiments (II) and (VI), the visual comparisons in Fig.~\ref{fig8} show that the regions highlighted by the boxes suffer from varying degrees of blurring, whereas the proposed model preserves smoother content transitions in the corresponding regions. Moreover, at the edge locations indicated by the triangles, both experiments exhibit conspicuous abrupt dark variations, suggesting the occurrence of temporal flickering. The quantitative results in Table~\ref{tab5} further show that experiments (II) and (VI) perform considerably worse than the proposed model across the overall metrics, demonstrating the importance of STIM and the temporal constraints.

For the ablations on pretrained feature extractors, namely experiments (III) and (IV), CLIP- and DINO-based variants achieve the best results on a few individual metrics. Nevertheless, their overall performance remains inferior to that of the proposed configuration using MAE, indicating that MAE features are more suitable for constructing the proposed temporal loss.

Finally, the comparison between experiment (V) and the proposed model shows that introducing $\mathcal{L}_{\mathrm{rev}}$ yields consistent improvements across the evaluation metrics, further validating its effectiveness in enhancing the overall performance of the proposed model.

\section{Limitation and Future Work}

The proposed method adopts a sequence-based fusion paradigm that enables effective sequence-level temporal interaction. In online inference, however, fixed-length clips must be buffered before joint processing, resulting in additional latency. Future work will investigate a streaming fusion framework with temporal state propagation, enabling incremental state updates to further reduce computational cost and improve inference speed.

\section{Conclusion}

This paper proposes an efficient MMVF algorithm that mainly relies on a TTT-based sequence modeling mechanism to effectively interact with dynamic regions along the temporal dimension. In addition, to suppress temporal flickering, we introduce a temporal consistency constraint that employs MAE as the feature extractor. The proposed framework not only produces high-quality fusion results on multiple public datasets, but also demonstrates higher inference efficiency when evaluated on an edge device. Finally, we introduce FD3 to evaluate the dynamic direction distance of temporal variations within dynamic regions in the pretrained MAE feature space.



 
%

\bibliographystyle{IEEEtran}
\bibliography{main}

@inproceedings{VITTT,
  title={{ViT}${}^3$: Unlocking Test-Time Training in Vision},
  author={Han, Dongchen and Li, Yining and Li, Tianyu and Cao, Zixuan and Wang, Ziming and Song, Jun and Cheng, Yu and Zheng, Bo and Huang, Gao},
  booktitle={Proceedings of the IEEE/CVF Conference on Computer Vision and Pattern Recognition},
  pages={51--61},
  year={2026}
}

@article{TIP1,
  title={Interpretable General Image Fusion via Scalable Autoregressive Modeling},
  author={Jingwei Xin and Boneng Shi and Zhen Liang and Junlin Hao and Xuehao Song and Nan Wang and Jie Li and Xinbo Gao},
  journal={IEEE Transactions on Image Processing},
  year={2026},
  volume={35},
  pages={6632-6644} 
}

@article{TIP2,
  title={Disentangle to fuse: Towards content preservation and cross-modality consistency for multi-modality image fusion},
  author={Qin, Xinran and Cui, Yuning and Sun, Shangquan and Chen, Ruoyu and Ren, Wenqi and Knoll, Alois and Cao, Xiaochun},
  journal={IEEE Transactions on Image Processing},
  year={2026},
  publisher={IEEE}
}

@article{lixilaitip2,
  title={Awm-fuse: Multi-modality image fusion for adverse weather via global and local text perception},
  author={Li, Xilai and Liu, Huichun and Li, Xiaosong and Ye, Tao and Kuang, Zhenyu and Li, Huafeng},
  journal={IEEE Transactions on Image Processing},
  year={2026},
  publisher={IEEE}
}

@article{TIP4,
  title={MDbFusion++: A Visible and Infrared Image Fusion Framework Capable for Motion Deblurring},
  author={Chen, Jun and Yu, Wei and Tian, Xin and Huang, Jun and Ma, Jiayi},
  journal={IEEE Transactions on Image Processing},
  volume={35},
  pages={3647--3661},
  year={2026},
  publisher={IEEE}
}

@article{TIP5,
  title={SDSFusion: A semantic-aware infrared and visible image fusion network for degraded scenes},
  author={Chen, Jun and Yang, Liling and Yu, Wei and Gong, Wenping and Cai, Zhanchuan and Ma, Jiayi},
  journal={IEEE Transactions on Image Processing},
  year={2025},
  publisher={IEEE}
}

@article{TIP6,
  title={Urfusion: Unsupervised unified degradation-robust image fusion network},
  author={Xu, Han and Yi, Xunpeng and Lu, Chen and Liu, Guangcan and Ma, Jiayi},
  journal={IEEE Transactions on Image Processing},
  year={2025},
  publisher={IEEE}
}

@article{lixilaitip,
  title={UMCFuse: A unified multiple complex scenes infrared and visible image fusion framework},
  author={Li, Xilai and Li, Xiaosong and Tan, Tianshu and Li, Huafeng and Ye, Tao},
  journal={IEEE Transactions on Image Processing},
  year={2025},
  publisher={IEEE}
}

@inproceedings{UPFusion,
  title={Text-guided channel perturbation and pre-trained knowledge integration for unified multi-modality image fusion},
  author={Li, Xilai and Li, Xiaosong and Jiang, Weijun},
  booktitle={Proceedings of the AAAI Conference on Artificial Intelligence},
  volume={40},
  number={8},
  pages={6521--6529},
  year={2026}
}

@article{tpamireview25,
  title={Infrared and visible image fusion: From data compatibility to task adaption},
  author={Liu, Jinyuan and Wu, Guanyao and Liu, Zhu and Wang, Di and Jiang, Zhiying and Ma, Long and Zhong, Wei and Fan, Xin and Liu, Risheng},
  journal={IEEE Transactions on Pattern Analysis and Machine Intelligence},
  volume={47},
  number={4},
  pages={2349--2369},
  year={2024},
  publisher={IEEE}
}

@article{mambavf,
  title={MambaVF: State Space Model for Efficient Video Fusion},
  author={Zhao, Zixiang and Cui, Yukun and Deng, Lilun and Bai, Haowen and Qin, Haotong and Feng, Tao and Schindler, Konrad},
  journal={arXiv preprint arXiv:2602.06017},
  year={2026}
}

@article{CoCoNet,
  title={Coconet: Coupled contrastive learning network with multi-level feature ensemble for multi-modality image fusion},
  author={Liu, Jinyuan and Lin, Runjia and Wu, Guanyao and Liu, Risheng and Luo, Zhongxuan and Fan, Xin},
  journal={International Journal of Computer Vision},
  volume={132},
  number={5},
  pages={1748--1775},
  year={2024},
  publisher={Springer}
}

@article{MaeFuse,
  title={MaeFuse: Transferring omni features with pretrained masked autoencoders for infrared and visible image fusion via guided training},
  author={Li, Jiayang and Jiang, Junjun and Liang, Pengwei and Ma, Jiayi and Nie, Liqiang},
  journal={IEEE Transactions on Image Processing},
  year={2025},
  publisher={IEEE}
}

@article{MAEDUNet,
  title={Learned Image Reasoning Prior Penetrates Deep Unfolding Network for Panchromatic and Multi-Spectral Image Fusion},
  author={Man Zhou and Jie Huang and Naishan Zheng and Chongyi Li},
  journal={2023 IEEE/CVF International Conference on Computer Vision (ICCV)},
  year={2023},
  pages={12364-12373}  
}

@article{QGQM,
  title={Objective Assessment of Multiresolution Image Fusion Algorithms for Context Enhancement in Night Vision: A Comparative Study},
  author={Zheng Liu and Erik Blasch and Zhiyun Xue and Jiying Zhao and Robert Lagani{\`e}re and Wei Wu},
  journal={IEEE Transactions on Pattern Analysis and Machine Intelligence},
  year={2012},
  volume={34},
  pages={94-109} 
}

@article{FSIM,
  title={FSIM: A Feature Similarity Index for Image Quality Assessment},
  author={Lin Zhang and Lei Zhang and Xuanqin Mou and David Zhang},
  journal={IEEE Transactions on Image Processing},
  year={2011},
  volume={20},
  pages={2378-2386}
}

@inproceedings{zhangjian1,
  title={Rethinking Occlusion Modeling for UAV Tracking},
  author={Zhang, Jian and Yu, Xincheng and Lin, Yi},
  booktitle={Proceedings of the IEEE/CVF Conference on Computer Vision and Pattern Recognition},
  pages={13563--13573},
  year={2026}
}

@article{xiaosongli1,
  title={Joint image fusion and denoising via three-layer decomposition and sparse representation},
  author={Xiaosong Li and Fuqiang Zhou and Haishu Tan},
  journal={Knowl. Based Syst.},
  year={2021},
  volume={224},
  pages={107087}
}

@article{ULMGFuse,
  title={An efficient image fusion network exploiting unifying language and mask guidance},
  author={Cao, Zi-Han and Liang, Yu-Jie and Deng, Liang-Jian and Vivone, Gemine},
  journal={IEEE Transactions on Pattern Analysis and Machine Intelligence},
  year={2025},
  publisher={IEEE}
}

@article{mask-dif,
  title={Mask-difuser: A masked diffusion model for unified unsupervised image fusion},
  author={Tang, Linfeng and Li, Chunyu and Ma, Jiayi},
  journal={IEEE Transactions on Pattern Analysis and Machine Intelligence},
  year={2025},
  publisher={IEEE}
}

@inproceedings{temcoco,
  title={Temcoco: Temporally consistent multi-modal video fusion with visual-semantic collaboration},
  author={Gong, Meiqi and Zhang, Hao and Yi, Xunpeng and Tang, Linfeng and Ma, Jiayi},
  booktitle={Proceedings of the IEEE/CVF International Conference on Computer Vision},
  pages={14326--14335},
  year={2025}
}

@article{VF-Bench,
  title={A unified solution to video fusion: From multi-frame learning to benchmarking},
  author={Zhao, Zixiang and Bai, Haowen and Ke, Bingxin and Cui, Yukun and Deng, Lilun and Zhang, Yulun and Zhang, Kai and Schindler, Konrad},
  journal={Advances in Neural Information Processing Systems},
  volume={38},
  pages={128504--128535},
  year={2026}
}

@article{DSPFusion,
  title={Dspfusion: Image fusion via degradation and semantic dual-prior guidance},
  author={Tang, Linfeng and Li, Chunyu and Wang, Yeda and Wang, Guoqing and Yuan, Yixuan and Ma, Jiayi},
  journal={IEEE Transactions on Image Processing},
  volume={35},
  pages={6331--6345},
  year={2026},
  publisher={IEEE}
}

@inproceedings{ISFusion,
  title={Multi-Modal Image Fusion via Intervention-Stable Feature Learning},
  author={Wang, Xue and Guan, Zheng and Qian, Wenhua and Wang, Chengchao and Ma, Runzhuo},
  booktitle = {Proceedings of the IEEE/CVF Conference on Computer Vision and Pattern Recognition (CVPR)},
  pages={33827--33837},
  year={2026}
}

@InProceedings{VideoFusion,
    author    = {Tang, Linfeng and Wang, Yeda and Gong, Meiqi and Li, Zizhuo and Deng, Yuxin and Yi, Xunpeng and Li, Chunyu and Xu, Han and Zhang, Hao and Ma, Jiayi},
    title     = {VideoFusion: A Spatio-Temporal Collaborative Network for Multi-modal Video Fusion},
    booktitle = {Proceedings of the IEEE/CVF Conference on Computer Vision and Pattern Recognition (CVPR)},
    month     = {June},
    year      = {2026},
    pages     = {19559-19569}
}

@article{HDO,
  title={Rcvs: A unified registration and fusion framework for video streams},
  author={Xie, Housheng and Sang, Meng and Zhang, Yukuan and Yang, Yang and Zhao, Shan and Zhong, Jianbo},
  journal={IEEE Transactions on Multimedia},
  volume={26},
  pages={11031--11043},
  year={2024},
  publisher={IEEE}
}

@article{VTUAV,
  title={Visible-Thermal UAV Tracking: A Large-Scale Benchmark and New Baseline},
  author={Pengyu Zhang and Jie Zhao and D. Wang and Huchuan Lu and Xiang Ruan},
  journal={2022 IEEE/CVF Conference on Computer Vision and Pattern Recognition (CVPR)},
  year={2022},
  pages={8876-8885} 
}

@article{SAMWISE,
  title={SAMWISE: Infusing Wisdom in SAM2 for Text-Driven Video Segmentation},
  author={Claudia Cuttano and Gabriele Trivigno and Gabriele Rosi and Carlo Masone and Giuseppe Averta},
  journal={2025 IEEE/CVF Conference on Computer Vision and Pattern Recognition (CVPR)},
  year={2025},
  pages={3395-3405}
}

@inproceedings{TTT,
  title={Learning to (Learn at Test Time): RNNs with Expressive Hidden States},
  author={Sun, Yu and Li, Xinhao and Dalal, Karan and Xu, Jiarui and Vikram, Arjun and Zhang, Genghan and Dubois, Yann and Chen, Xinlei and Wang, Xiaolong and Koyejo, Sanmi and others},
  booktitle={International Conference on Machine Learning},
  pages={57503--57522},
  year={2025},
  organization={PMLR}
}

@article{VIT,
  title={Attention is all you need},
  author={Vaswani, Ashish and Shazeer, Noam and Parmar, Niki and Uszkoreit, Jakob and Jones, Llion and Gomez, Aidan N and Kaiser, {\L}ukasz and Polosukhin, Illia},
  journal={Advances in neural information processing systems},
  volume={30},
  year={2017}
}

@inproceedings{linearattention,
  title={Transformers are rnns: Fast autoregressive transformers with linear attention},
  author={Katharopoulos, Angelos and Vyas, Apoorv and Pappas, Nikolaos and Fleuret, Fran{\c{c}}ois},
  booktitle={International Conference on Machine Learning},
  pages={5156--5165},
  year={2020},
  organization={PMLR}
}

@inproceedings{han2023flatten,
  title={Flatten transformer: Vision transformer using focused linear attention},
  author={Han, Dongchen and Pan, Xuran and Han, Yizeng and Song, Shiji and Huang, Gao},
  booktitle={Proceedings of the IEEE/CVF International Conference on Computer Vision},
  pages={5961--5971},
  year={2023}
}

@inproceedings{qincosformer,
  title={cosFormer: Rethinking Softmax In Attention},
  author={Qin, Zhen and Sun, Weixuan and Deng, Hui and Li, Dongxu and Wei, Yunshen and Lv, Baohong and Yan, Junjie and Kong, Lingpeng and Zhong, Yiran},
  booktitle={International Conference on Learning Representations}
}

@inproceedings{SEBlock,
  title={Squeeze-and-excitation networks},
  author={Hu, Jie and Shen, Li and Sun, Gang},
  booktitle={Proceedings of the IEEE Conference on Computer Vision and Pattern Recognition},
  pages={7132--7141},
  year={2018}
}

@article{DINOv2,
  title={DINOv2: Learning Robust Visual Features without Supervision},
  author={Oquab, Maxime and Darcet, Timoth{\'e}e and Moutakanni, Th{\'e}o and Vo, Huy and Szafraniec, Marc and Khalidov, Vasil and Fernandez, Pierre and Haziza, Daniel and Massa, Francisco and El-Nouby, Alaaeldin and others},
  journal={Transactions on Machine Learning Research Journal},
  year={2024}
}

@inproceedings{CLIP,
  title={Learning transferable visual models from natural language supervision},
  author={Radford, Alec and Kim, Jong Wook and Hallacy, Chris and Ramesh, Aditya and Goh, Gabriel and Agarwal, Sandhini and Sastry, Girish and Askell, Amanda and Mishkin, Pamela and Clark, Jack and others},
  booktitle={International Conference on Machine Learning},
  pages={8748--8763},
  year={2021},
  organization={PmLR}
}

@inproceedings{GIFNet,
  title={One Model for ALL: Low-Level Task Interaction Is a Key to Task-Agnostic Image Fusion},
  author={Cheng, Chunyang and Xu, Tianyang and Feng, Zhenhua and Wu, Xiaojun and Tang, Zhangyong and Li, Hui and Zhang, Zeyang and Atito, Sara and Awais, Muhammad and Kittler, Josef},
  booktitle={Proceedings of the Computer Vision and Pattern Recognition Conference},
  pages={28102--28112},
  year={2025}
}

@article{Huangjingxue1,
  title={Multiple attention channels aggregated network for multimodal medical image fusion},
  author={Jingxue Huang and Tianshu Tan and Xiaosong Li and Tao Ye and Yanxiong Wu},
  journal={Medical Physics},
  year={2024},
  volume={52},
  pages={2356-2374}
}

@article{DeDNet,
  title={DeDNet: Infrared and visible image fusion with noise removal by decomposition-driven network},
  author={Huang, Jingxue and Li, Xiaosong and Tan, Haishu and Yang, Lemiao and Wang, Gao and Yi, Peng},
  journal={Measurement},
  pages={115092},
  year={2024},
  publisher={Elsevier}
}

@article{wavefusion,
  title={WaveFusion: A novel wavelet vision transformer with saliency-guided enhancement for multimodal image fusion},
  author={Wang, Qinghua and Li, Ziwei and Zhang, Shuqi and Chi, Nan and Dai, Qionghai},
  journal={IEEE Transactions on Circuits and Systems for Video Technology},
  year={2025},
  publisher={IEEE}
}

@article{mma,
  title={Mma-unet: A multi-modal asymmetric unet architecture for infrared and visible image fusion},
  author={Huang, Jingxue and Li, Xilai and Tan, Tianshu and Li, Xiaosong and Zhou, Fuqiang and Li, Huafeng},
  journal={IEEE Transactions on Emerging Topics in Computational Intelligence},
  year={2026},
  publisher={IEEE}
}

@article{EvaNet,
  title={EvaNet: Towards More Efficient and Consistent Infrared and Visible Image Fusion Assessment},
  author={Cheng, Chunyang and Xu, Tianyang and Wu, Xiao-Jun and Zhou, Tao and Li, Hui and Tang, Zhangyong and Kittler, Josef},
  journal={IEEE Transactions on Pattern Analysis and Machine Intelligence},
  year={2026},
  publisher={IEEE}
}

@article{LRRNet,
  title={LRRNet: A novel representation learning guided fusion network for infrared and visible images},
  author={Li, Hui and Xu, Tianyang and Wu, Xiao-Jun and Lu, Jiwen and Kittler, Josef},
  journal={IEEE Transactions on Pattern Analysis and Machine Intelligence},
  volume={45},
  number={9},
  pages={11040--11052},
  year={2023},
  publisher={IEEE}
}

@inproceedings{MAE,
  title={Masked autoencoders are scalable vision learners},
  author={He, Kaiming and Chen, Xinlei and Xie, Saining and Li, Yanghao and Doll{\'a}r, Piotr and Girshick, Ross},
  booktitle={Proceedings of the IEEE/CVF Conference on Computer Vision and Pattern Recognition},
  pages={16000--16009},
  year={2022}
}

@article{shiyuCVPR26,
  title={Degradation-Robust Fusion: An Efficient Degradation-Aware Diffusion Framework for Multimodal Image Fusion in Arbitrary Degradation Scenarios},
  author={Shi, Yu and Liu, Yu and Wu, Zhong-Cheng and Cheng, Juan and Li, Huafeng and Chen, Xun},
  journal={arXiv preprint arXiv:2604.08922},
  year={2026}
}

@article{shufflemamba,
  title={Shuffle mamba: State space models with random shuffle for multi-modal image fusion},
  author={Cao, Ke and He, Xuanhua and Hu, Tao and Xie, Chengjun and Zhou, Man and Zhang, Jie},
  journal={IEEE Transactions on Circuits and Systems for Video Technology},
  year={2026},
  publisher={IEEE}
}

@article{PRRGAN,
  title={Generative adversarial network for trimodal medical image fusion using primitive relationship reasoning},
  author={Huang, Jingxue and Li, Xiaosong and Tan, Haishu and Cheng, Xiaoqi},
  journal={IEEE Journal of Biomedical and Health Informatics},
  volume={28},
  number={10},
  pages={5729--5741},
  year={2024},
  publisher={IEEE}
}

@article{xia2026rethinking,
  title={Rethinking the CNN-Transformer hybrid architecture for infrared and visible image fusion from a frequency perspective},
  author={Xia, Jianghan and Song, Hong and Li, Jinfu and Ma, Shihan and Lin, Yucong and Huang, Yuqi and Yang, Jian},
  journal={IEEE Transactions on Multimedia},
  year={2026},
  publisher={IEEE}
}

@article{Transformer1,
  title={Fully-connected transformer for multi-source image fusion},
  author={Wu, Xiao and Cao, Zi Han and Huang, Ting Zhu and Deng, Liang Jian and Chanussot, Jocelyn and Vivone, Gemine},
  journal={IEEE Transactions on Pattern Analysis and Machine Intelligence},
  volume={47},
  number={3},
  pages={2071--2088},
  year={2025},
  publisher={IEEE}
}

@article{Frefusion,
  title={Frefusion: Frequency domain transformer for infrared and visible image fusion},
  author={Shi, Junjie and Duan, Puhong and Ma, Xiaoguang and Chi, Jianning and Dai, Yong},
  journal={IEEE Transactions on Multimedia},
  year={2025},
  publisher={IEEE}
}

@article{DDCGAN,
  title={DDcGAN: A dual-discriminator conditional generative adversarial network for multi-resolution image fusion},
  author={Ma, Jiayi and Xu, Han and Jiang, Junjun and Mei, Xiaoguang and Zhang, Xiao-Ping},
  journal={IEEE Transactions on Image Processing},
  volume={29},
  pages={4980--4995},
  year={2020},
  publisher={IEEE}
}

@article{GAN1,
  title={Transformer based conditional GAN for multimodal image fusion},
  author={Zhang, Jun and Jiao, Licheng and Ma, Wenping and Liu, Fang and Liu, Xu and Li, Lingling and Chen, Puhua and Yang, Shuyuan},
  journal={IEEE Transactions on Multimedia},
  volume={25},
  pages={8988--9001},
  year={2023},
  publisher={IEEE}
}

@article{mamba1,
  title={Mamba collaborative implicit neural representation for hyperspectral and multispectral remote sensing image fusion},
  author={Zhu, Chunyu and Deng, Shangqi and Song, Xuan and Li, Yachao and Wang, Qi},
  journal={IEEE Transactions on Geoscience and Remote Sensing},
  year={2025},
  publisher={IEEE}
}

@article{mamba2,
  title={HCFMaNet: A novel holistic cross-modal fusion mamba network for multi-modal medical image fusion},
  author={Wei, Xinjian and Qiu, Yu and Xu, Jing and Zhang, Jun},
  journal={IEEE Transactions on Multimedia},
  year={2026},
  publisher={IEEE}
}

@article{CNN1,
  title={Image Fusion Based on Prior Information.},
  author={Congbing He and Zhenhong Jia and Hui Zhao and Sensen Song and Jiajia Wang and Fei Shi},
  journal={IEEE Transactions on Pattern Analysis and Machine Intelligence},
  year={2026},
  volume={PP}
}

@article{Diffusion1,
  title={Efficient rectified flow for image fusion},
  author={Wang, Zirui and Zhang, Jiayi and Guan, Tianwei and Zhou, Yuhan and Li, Xingyuan and Dong, Minjing and Liu, Jinyuan},
  journal={Advances in Neural Information Processing Systems},
  volume={38},
  pages={22927--22948},
  year={2026}
}











\newpage

 




\vfill

\end{document}